\pdfoutput=1

\documentclass[11pt]{article}
\usepackage[final]{acl}
\usepackage[dvipsnames]{xcolor}
\usepackage{times}
\usepackage{latexsym}
\usepackage{placeins}
\usepackage{rotating}
\usepackage{amsmath}
\usepackage{graphicx}
\usepackage[most]{tcolorbox}
\usepackage{amssymb}
\usepackage[T1]{fontenc}
\usepackage{lipsum}

\usepackage[utf8]{inputenc}

\usepackage{microtype}
\usepackage{enumitem}
\usepackage{inconsolata}

\usepackage{graphicx}

\title{Indic-TunedLens: Interpreting Multilingual Models in Indian Languages}

\author{
Mihir Panchal$^{1}$, Deeksha Varshney$^{2}$, Mamta .$^{3}$, Asif Ekbal$^{4}$ \\
$^{1}$ Dwarkadas Jivanlal Sanghvi College of Engineering, Mumbai, India \\
$^{2}$ Indian Institute of Technology Jodhpur, Jodhpur, India \\
$^{3}$ King’s College London, London, UK \\
$^{4}$ Indian Institute of Technology Patna, Patna, India \\
\texttt{mihirpanchal5400@gmail.com, deeksha@iitj.ac.in, } \\
\texttt{mamta.name@kcl.ac.uk, asif@iitp.ac.in}
}

\begin{document}
\maketitle
\begin{abstract}

Multilingual large language models (LLMs) are increasingly deployed in linguistically diverse regions like India, yet most interpretability tools remain tailored to English. Prior work reveals that LLMs often operate in English centric representation spaces, making cross lingual interpretability a pressing concern. We introduce \textit{Indic-TunedLens}, a novel interpretability framework specifically for Indian languages that learns shared affine transformations. Unlike the standard Logit Lens, which directly decodes intermediate activations, \textit{Indic-TunedLens} adjusts hidden states for each target language, aligning them with the target output distributions to enable more faithful decoding of model representations. We evaluate our framework on 10 Indian languages using the MMLU benchmark and find that it significantly improves over SOTA interpretability methods, especially for morphologically rich, low resource languages. Our results provide crucial insights into the layer-wise semantic encoding of multilingual transformers. Our model is available at \url{https://huggingface.co/spaces/MihirRajeshPanchal/IndicTunedLens}. Our code is available at \url{https://github.com/MihirRajeshPanchal/IndicTunedLens}.

\end{abstract}

\section{Introduction}

The remarkable advancements of transformer models \cite{vaswani2017attention} across diverse domains, particularly in Natural Language Processing (NLP), underscore the critical need to decipher their internal representations and intricate reasoning processes.
This is essential to ensure trust and reliability in the systems.
Existing research has explored various methods to extract specific linguistic concepts, such as speech tagging parts or syntactic structure, from hidden states with the help of probing classifiers \cite{li2023emergent,lee2025geometry} and LogitLens \cite{nostalgebraist2020logitlens}. \citet{belrose2023eliciting} interpret the layers which offered a significant step forward in extracting meaningful intermediate predictions by training layer specific translators. Their approach builds upon the concept of {LogitLens}, which directly decodes hidden states into the vocabulary space using the model’s pre-trained unembedding matrix. Most of these approaches fail to generalize effectively to linguistically diverse and low resource languages.

India is a multilingual country and Indian languages exhibit vastly different linguistic structures from English, including rich morphology and more flexible word order (common in Hindi and Punjabi) \cite{srirampur-etal-2014-statistical}. These diverse scripts (e.g., Gurmukhi for Punjabi, Devanagari for Hindi and Marathi) and their unique tokenization schemes which often result in distinct subword units and vocabulary distributions, pose another significant challenges in applying English interpretability models to Indian languages \cite{rahman2024indicllmsuite}. \citet{schut2025multilingual} highlights a critical limitation of current multilingual LLMs, even when prompted and generating outputs in non-English languages, the models tend to operate in an English centric latent space. Using logit lens decoding, they show that LLMs first activate English word representations before translating them into the target language, suggesting that key semantic reasoning is biased toward English. However, this analysis is limited to European and East Asian languages, and does not explore morphologically rich and typologically distinct languages such as those found in the Indian subcontinent.

To address these critical limitations, we developed \textit{Indic-TunedLens} framework specifically for multilingual contexts, and demonstrate its efficacy on transformer models for 10 low resource indian languages. Our approach involves training a single affine transformation to learn shared representations across languages, while also enabling alignment of intermediate hidden states with the final output distribution for Indic languages. This allows the lens to capture the unique representational nuances of these languages, providing robust and interpretable insights into how these diverse and morphologically rich languages are processed layer by layer within transformer models. 
We conduct a comprehensive evaluation on the Sarvam-1 model \footnote{https://www.sarvam.ai/blogs/sarvam-1} using Bengali, English, Gujarati, Hindi, Kannada, Malayalam,  Marathi, Panjabi, Nepali, Tamil and Telugu languages with a multilingual MMLU benchmark \cite{dac2023okapi} to demonstrate the effectiveness of \textit{Indic-TunedLens} framework.
Our findings not only bridge a significant gap in multilingual transformer interpretability but also offer a crucial tool for understanding and enhancing NLP applications across the rich linguistic tapestry of India. To the best of our knowledge, this is the first work to apply layer wise interpretability to multilingual LLMs for Indian languages. Our work makes the following contributions:

\begin{enumerate} [noitemsep]
    \item We show that interpretability failure in Indian languages is a projection problem, and propose \textit{Indic-TunedLens }as a solution. It aligns intermediate hidden states with output distributions in morphologically rich and syntactically diverse low resource indic languages.
    \item We conduct a comprehensive evaluation on the Sarvam-1 model on Bengali, English, Gujarati, Hindi, Kannada, Malayalam,  Marathi, Nepali, Tamil and Telugu languages with a multilingual MMLU benchmark. 
    \item Our results show that \textit{Indic-TunedLens} enables accurate decoding of intermediate representations and reveals distinct layer-wise and token position specific reasoning dynamics across languages.
\end{enumerate}

\section{Related Work}

\subsection{Interpretability in Transformer Models}
The growing complexity of transformer based language models has led to increased interest in interpretability techniques that help unpack how internal representations evolve across layers. Early efforts, such as probing classifiers \cite{li2023emergent, lee2025geometry} attempted to extract linguistic features like syntax and part-of-speech information from hidden states. Methods, like the {Logit Lens} \cite{nostalgebraist2020logitlens}, directly decode intermediate activations into vocabulary space using the model’s unembedding matrix. Building on this, the {Tuned Lens} \cite{belrose2023eliciting} improves interpretability by learning affine transformations that align hidden states with the final output distribution. Complementary methods, such as causal tracing \cite{meng2022locating} and representation steering \cite{subramani2022directional} have also enabled insights into factual recall and controlled generation.

\subsection{Multilingual Interpretability}
India's linguistic diversity presents unique challenges for model interpretability. \citet{wen2023hyperpolyglot} demonstrate that multilingual models learn input embeddings where translated tokens across languages cluster together, even without an explicit translation objective. Similarly, \citet{gopinath2024probing} observe that self-supervised speech models exhibit diverse attention head behaviors across languages, with diagonal heads playing a crucial role in cross lingual phoneme classification. \citet{schut2025multilingual} demonstrate that multilingual LLMs often process semantically meaningful content in an English centric representation space, regardless of input/output language. This bias becomes more pronounced for lexical tokens, affecting model transparency and fairness. Despite these advances, most interpretability research remains centered on English or non-LLM architectures. Notably, \citet{siddiqui2024fine} use LIME to interpret hate speech classifiers in low-resource languages like Urdu and Sindhi, however, this work focuses on traditional classification tasks, not on large language models or multilingual reasoning. Additionally, \citet{saji2025romanlens} show that LLMs often represent non-Roman script languages in Romanized form in intermediate layers, a phenomenon termed Latent Romanization highlighting shared cross-script representations that complement prior findings on English-centric biases.

In contrast to these prior approaches, we propose \textit{Indic-TunedLens}, an interpretability framework tailored for Indian languages. Unlike existing methods that apply English-tuned lens, our approach learns a shared affine transformation adapted to Indic scripts, aligning hidden states with vocabulary distributions for Bengali, English, Gujarati, Hindi, Kannada, Malayalam,  Marathi, Panjabi, Nepali, Tamil and Telugu, and enabling fine-grained, language-aware analysis of transformer model representations.

\section{Indic-TunedLens}

The {Tuned Lens} was introduced by \cite{nostalgebraist2020logitlens}, is a probing technique designed to interpret intermediate hidden states of transformer models by projecting them into the final output space. It enables inspection of the model's predictions at each layer without requiring task specific supervision. 

We adapt this framework to Indian languages and introduce \textit{Indic-TunedLens}, which applies the same affine transformation approach to better capture the unique challenges posed by Indic scripts and morphologically rich, low-resource settings. Unlike existing applications that evaluate Tuned Lens in English-centric contexts, our method explicitly trains and evaluates on Indian languages, providing new insights into their representation dynamics.

\paragraph{Affine Translation of Representations.} For a transformer layer $n$, let $h_n$ denote the hidden state. The \textit{Indic-TunedLens} maps $h_n$ to a distribution over vocabulary tokens by first applying a learned affine transformation, a \textit{translator}, comprising a matrix $M_n \in \mathbb{R}^{d \times d}$ and a bias vector $b_n \in \mathbb{R}^d$. The transformed representation is then passed through the model's final layer head to produce token logits:
\begin{equation}
\begin{split}
\text{Indic-TunedLens}_n(h_n) &= \text{LogitHead}(\tilde{h}_n) \\
\text{where } \tilde{h}_n &= M_n h_n + b_n
\end{split}
\end{equation}
where \texttt{LogitHead} refers to the model’s output projection layer (typically a linear map followed by softmax over the vocabulary).

\paragraph{Learning Objective.} The \textit{Indic-TunedLens} is trained to reproduce the predictions of the base model’s final layer. For a given input $x$, the base model produces a next-token probability distribution $p_{\text{final}}(x)$ at its output layer. We treat this distribution as the supervision signal. For each intermediate layer $n$, the \textit{Indic-TunedLens} produces its own distribution by translating the hidden state $h_n$ and passing it through the model’s output head. We minimize the Kullback-Leibler divergence between these two distributions:

\begin{equation}
\begin{split}
\min_{M_n, b_n} \ \mathbb{E}_{h_n} \Big[ &D_{\text{KL}}\big(p_{\text{final}}(x) \,\|\, \\
&\text{Indic-TunedLens}_n(h_n) \big) \Big]
\end{split}
\end{equation}

where $p_{\text{final}}(x) = \text{softmax}(W h_L)$ denotes the next-token probability distribution produced by the model's final layer $L$ on input $x$, and $h_n$ is the hidden state at intermediate layer $n$.

In other words, the model’s own final-layer predictions serve as the ``labels,'' and the \textit{Indic-TunedLens} learns to align intermediate hidden states with this distribution.

\section{Experimental Setup and Analysis}
\label{sec:exp_setup}

\textit{Indic-TunedLens} is trained using the Sarvam-1 model as the base model.
All the implementations are carried out in Pytorch \footnote{https://pytorch.org/}.
To train and test \textit{Indic-TunedLens}, we utilize two datasets that provide comprehensive multilingual coverage across Indian languages. We train our model on eleven languages,  Bengali, English, Gujarati, Hindi, Kannada, Malayalam,  Marathi, Panjabi, Nepali, Tamil and Telugu from the Sangraha dataset \citep{rahman2024indicllmsuite}.
For evaluation, we employ a curated subset of the multilingual Massive Multitask Language Understanding (MMLU) dataset \citep{dac2023okapi} adapted for Indian languages, comprising Bengali, English, Gujarati, Hindi, Kannada, Malayalam,  Marathi, Nepali, Tamil and Telugu samples. The inclusion of Panjabi as a train only language enables assessment of cross lingual transfer capabilities to unseen but related languages. 
Detailed dataset statistics and implementation details are provided in Appendices \ref{appendix:sangraha} and \ref{appendix:implementationdetails}, respectively.

\paragraph{Entropy Analysis.}\label{entropy} To analyze the confidence and uncertainty of predictions at each layer, we compute the Shannon entropy of the probability distribution produced by the \textit{Indic-TunedLens}:
\begin{equation}
H_n = -\sum_{i=1}^{|V|} p_i^{(n)} \log p_i^{(n)},
\end{equation}
where $p_i^{(n)}$ is the probability assigned to vocabulary token $i$ at layer $n$, and $|V|$ is the vocabulary size. Lower entropy values indicate confident predictions (probability mass concentrated on fewer tokens), while higher entropy values indicate uncertain predictions (probability mass distributed across many tokens).

\begin{figure*}
    \centering
    \includegraphics[width=1\linewidth]{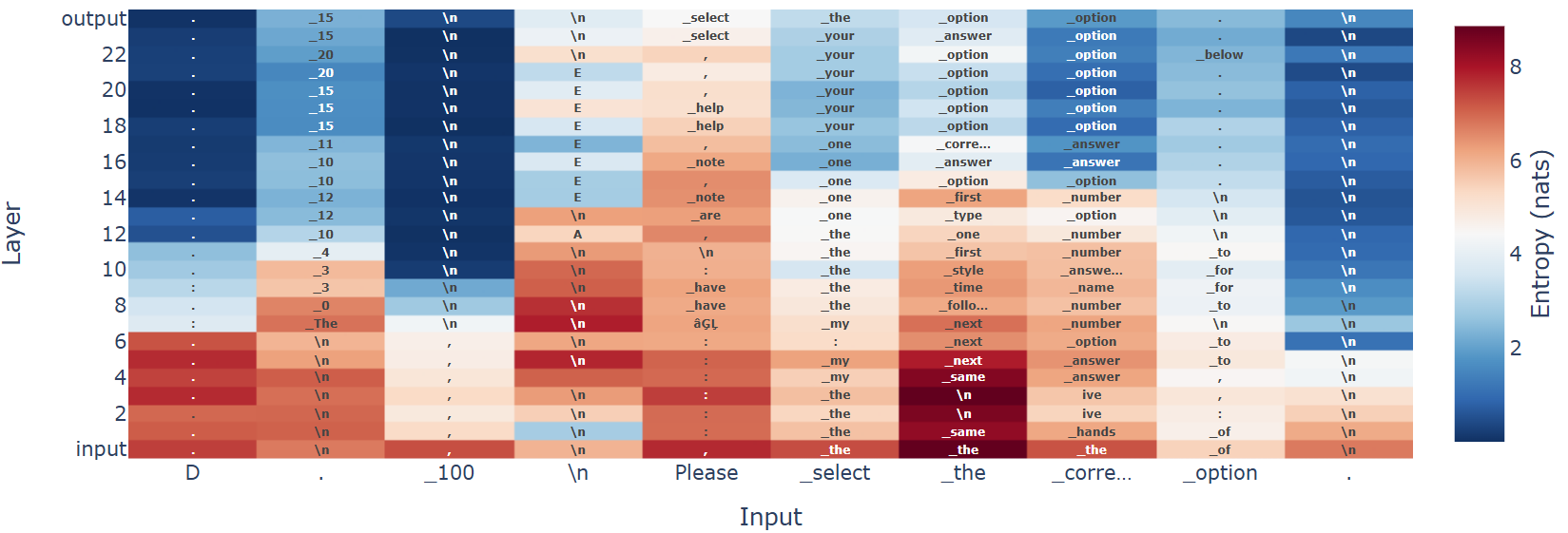}
    \caption{This figure shows the entropy heatmap of the standard Tuned Lens, which was developed for English-centric models. The high and irregular entropy across layers suggests unstable intermediate representations and weak alignment for Indian languages, with predictions biased toward English tokens.}
    \label{fig:tuned-lens}
\end{figure*}

\begin{figure*}[t!]
    \centering
    \includegraphics[width=1\linewidth]{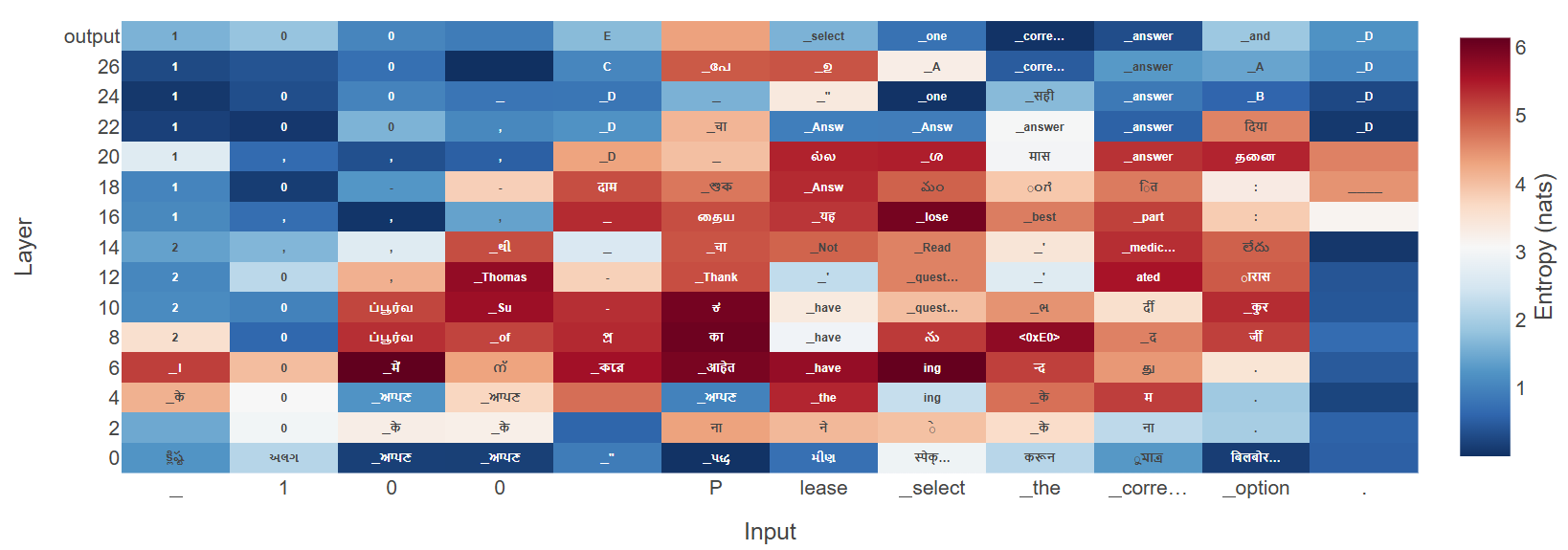}
    \caption{Entropy heatmap for the Indic-TunedLens. Entropy decreases more smoothly across layers, indicating progressive information consolidation and improved semantic alignment, with intermediate predictions increasingly generating meaningful Hindi tokens.}
    \label{fig:indictunedlens}
\end{figure*}

\paragraph{Layer-wise Accuracy}\label{layerwise_accuracy_def}
We quantify alignment between intermediate representations and the final next-token prediction using a layer-wise agreement score. For a sequence of length $T$, the top-1 token predicted by a lens at layer $n$ for position $t$ is
\begin{equation}
\hat{y}_t^{(n)} = \arg\max_{i \in V} \, p_i^{(n)}(\cdot \mid t).
\end{equation}
Agreement is measured against the base model’s prediction at the final layer $L$. This interpretability metric indicates how early sufficient information for the final decision emerges. Higher early-layer agreement reflects better alignment of intermediate states with the vocabulary, enabling faithful intermediate predictions.


\section{Results}

To assess the effectiveness of Indic-TunedLens for multilingual transformer interpretability, we conduct a comprehensive comparative analysis against the standard Logit Lens approach across 10 Indian languages (detailed in Appendix \ref{experiment_setup}).

\begin{figure}[t!]
    \centering
    \includegraphics[width=1\linewidth]{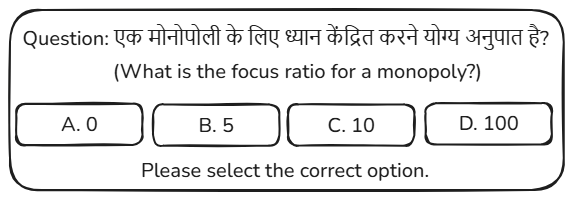}
    \caption{Input Question from MMLU Hindi Dataset}
    \label{fig:question}
\end{figure}

\begin{figure*}[ht!]
    \centering
    \includegraphics[width=0.90\linewidth]{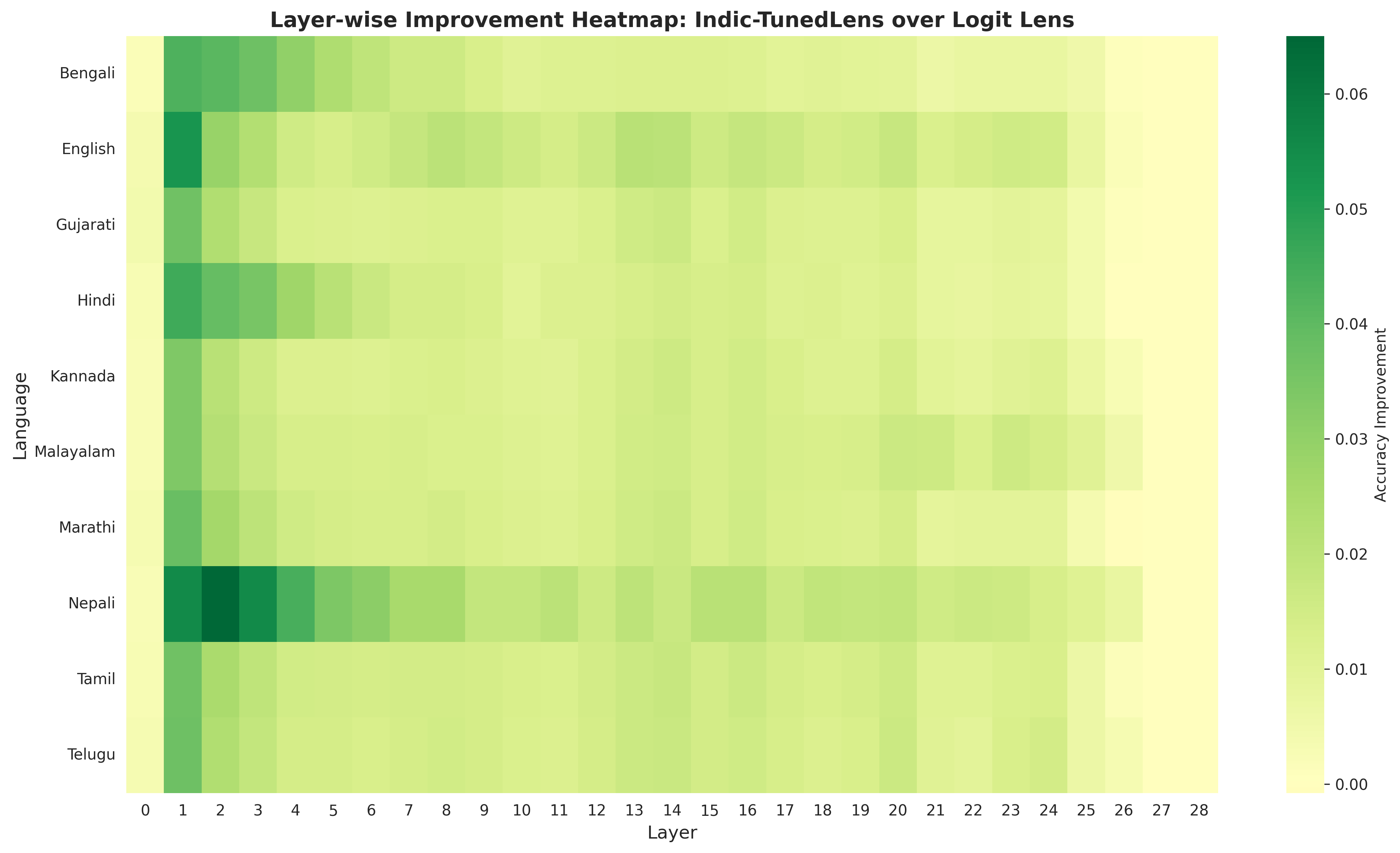}
    \caption{Layer-wise Improvement Patterns}
    \label{fig:layer_heatmap}
\end{figure*}

\noindent \textbf{Layer-wise uncertainty reveals language-specific meaning formation}. The standard Tuned Lens approach, originally designed for English centric models, demonstrates significant limitations when applied to Indian language processing, as illustrated in Figure \ref{fig:tuned-lens}. We show an example question (Figure \ref{fig:question}) about monopoly concentration ratios from the Hindi MMLU dataset\footnote{https://huggingface.co/datasets/alexandrainst/m\_mmlu/
viewer/hi}. The entropy (calculated as discussed in Section \ref{sec:exp_setup}) heatmap which measures uncertainty in the model’s predicted token distribution at each layer reveals inconsistent patterns with high variance across layers, indicating poor alignment between the tuning mechanism and the model's internal processing pathways for Indian languages. The visualization shows fragmented attention patterns, particularly in the middle layers (8-16), where semantic consolidation should occur, and scattered high entropy regions without clear progressive information refinement. This degraded performance stems from a projection mismatch: while the model internally encodes morphologically rich and semantically meaningful representations for Indian languages, the English-centric Tuned Lens fails to align these representations with the output vocabulary space.

The Indic-Tuned Lens, trained on Indian languages, demonstrates substantially improved performance characteristics as shown in Figure \ref{fig:indictunedlens}. The entropy heatmap exhibits coherent, systematic entropy reduction patterns across layers, indicating proper information consolidation aligned with Indic linguistic structures. Enhanced semantic capture is evident through concentrated attention patterns around key vocabulary elements, while distinct processing phases are observable with early layers managing tokenization and morphological analysis, middle layers handling semantic understanding, and later layers focusing on answer. Without language-aware projection, intermediate representations in Indian languages appear fragmented and difficult to interpret, obscuring how meaning forms across layers. The superior performance of \textit{Indic-TunedLens} demonstrates the critical need for developing specialized interpretability frameworks that account for diverse linguistic structures, and semantic processing patterns beyond English centric approaches.

\begin{figure*}[ht!]
    \centering
    \includegraphics[width=1\linewidth]{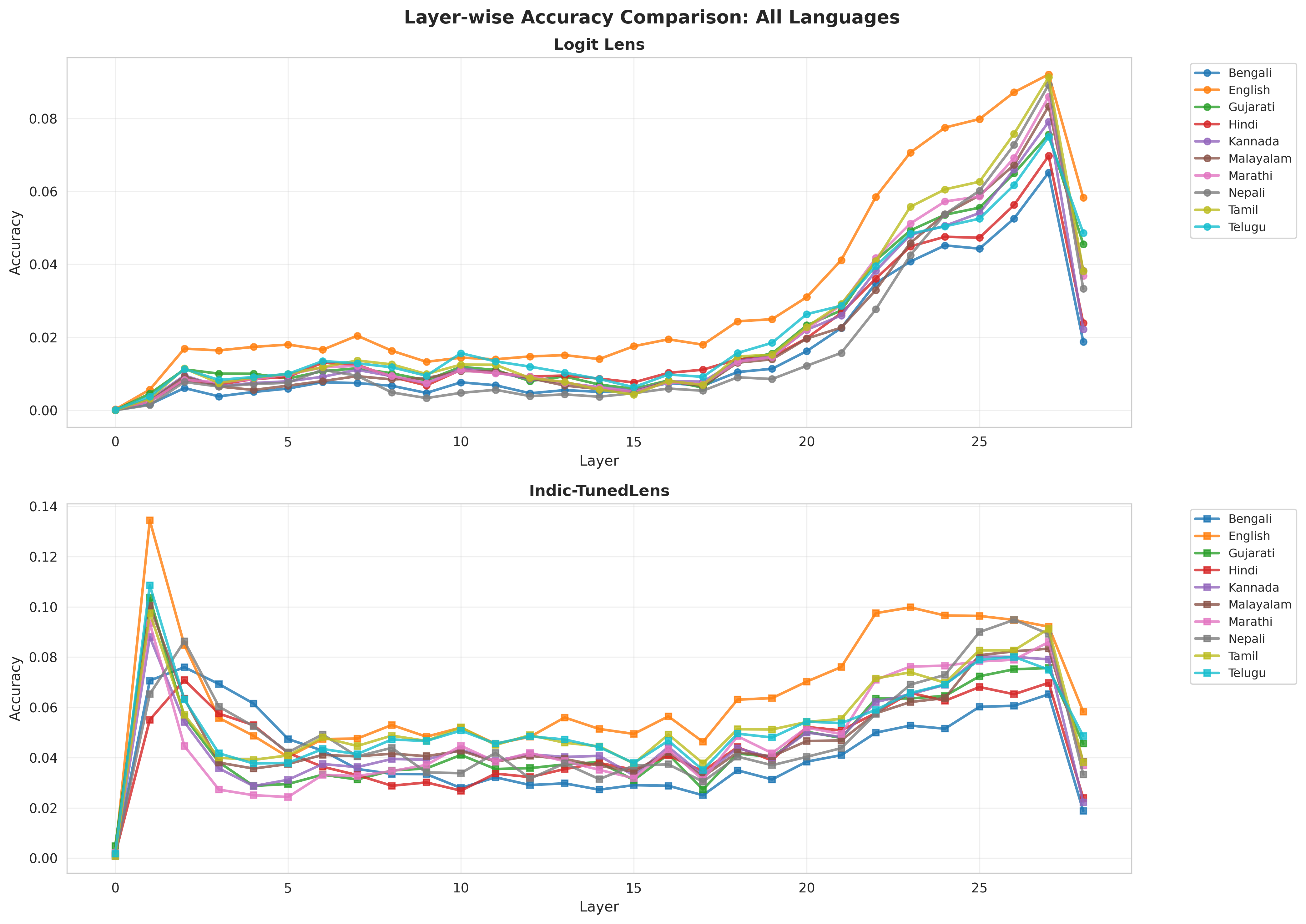}
    \caption{Layer Wise Accuracy Comparison}
    \label{fig:layer_comp}
\end{figure*}

\noindent \textbf{Maximum improvements occur in early layers (1-8) with language specific patterns reflecting morphological complexity.} The heatmap in Figure \ref{fig:layer_heatmap} shows that Bengali, Nepali, and Hindi exhibit the strongest improvements in layers 1-4 (0.04-0.06 accuracy gain), corresponding to the morphological analysis phase where rich inflectional systems are processed. English shows concentrated improvements at layer 1 (0.06 gain) and more modest gains throughout middle layers, reflecting its analytic morphology. Gujarati and Kannada show very similar sustained improvements across layers 1-10 (0.02-0.04 gain), while Malayalam exhibits more distributed improvement patterns extending into middle layers (5-15). {These layer specific improvement patterns enable linguistic hypotheses about where different linguistic phenomena are computed, suggesting that morphological analysis and early semantic composition occur in early layers, while deeper layers support more complex compositional and syntactic integration. This process happens earlier in languages like Telugu and Tamil, which form words by joining many meaningful parts together, compared to languages with more isolated word structures. The variation in improvement patterns across languages suggests that the learned transformations adapt to language specific processing dynamics, capturing the distinct computational trajectories by which different Indic languages resolve ambiguity and build compositional meaning.} These findings demonstrate that interpretability methods cannot be language-agnostic: effective analysis of multilingual LLMs requires language-family-aware lenses that respect typological differences in morphological and compositional structure.

\noindent \textbf{Indic-TunedLens demonstrates superior accuracy over Logit Lens across all Indian languages, with particularly pronounced improvements in early and middle layers.} As shown in Figure \ref{fig:layer_comp}, the standard Logit Lens exhibits minimal accuracy in early layers (0-15) across all languages, with values remaining below 0.02 for most Indian languages. The accuracy only begins to increase substantially after layer 20, reaching peak performance at layer 27 before dropping sharply at the final layer. In contrast, Indic-TunedLens shows immediate interpretability from layer 1, with English achieving 0.13 accuracy, followed by Telugu (0.11) and Gujarati (0.10). The framework maintains consistent accuracy between 0.03-0.06 throughout middle layers (5-20) for Indian languages, demonstrating robust intermediate representation capture. {These findings suggest that language specific affine transformations enable meaningful interpretation of hidden states from the earliest layers of processing, whereas English centric approaches fail to capture the morphological richness and syntactic diversity inherent in Indian languages until much deeper in the network. The superior early layer performance of Indic-TunedLens indicates that specialized training on diverse Indic scripts creates representations better aligned with the model's internal processing pathways for these languages, enabling more accurate decoding of intermediate semantic states. Individual language specific accuracy curves across all layers are provided in the appendix \ref{da: language_acc}}.

\begin{figure*}[t!]
    \centering
    \includegraphics[width=1\linewidth]{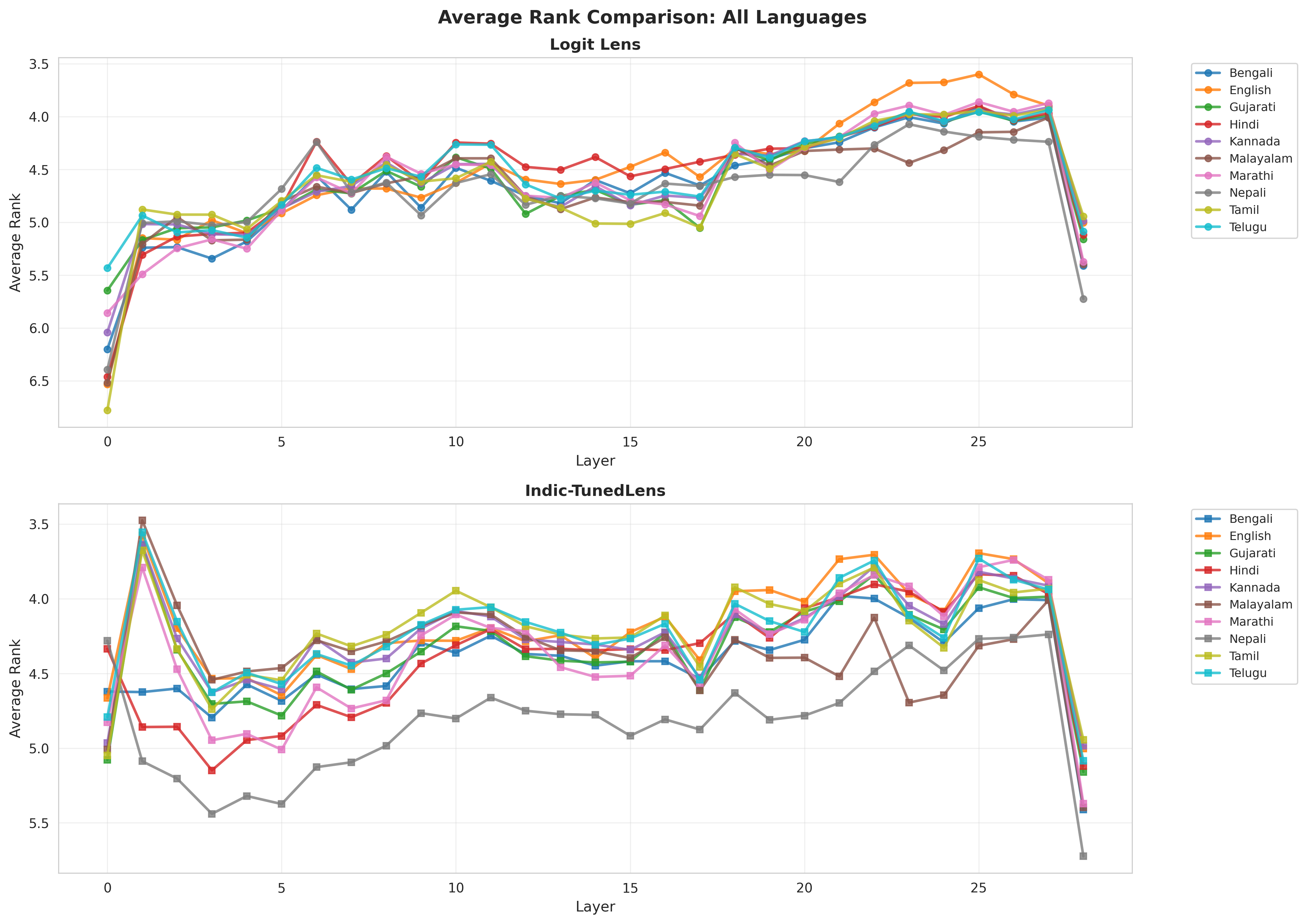}
    \caption{Average Rank of Correct Predictions}
    \label{fig:avg_rank}
\end{figure*}

\noindent \textbf{Indic-TunedLens consistently positions correct tokens at higher ranks across all layers, indicating improved confidence in predictions.} Figure \ref{fig:avg_rank} reveals that Logit Lens maintains average ranks between 4-5 across layers 5-25 for most languages, with Malayalam showing particularly poor performance (ranks 4.5-5.8). The ranks only improve dramatically at layer 27, converging to 3.5-4 across languages. Conversely, Indic-TunedLens exhibits more favorable ranking patterns throughout the network, with most languages maintaining average ranks between 3.9-4.7 across middle layers. Notably, Hindi demonstrates superior rank positioning (4.5-5 in early layers, improving to 3.5-4.0 in later layers), while Malayalam shows the most substantial improvement from its Logit Lens baseline, achieving ranks of 4.2-4.8 in layers 10-25. {The consistently better ranking performance of Indic-TunedLens across all languages demonstrates that the learned affine transformations more effectively align intermediate representations with the final vocabulary distribution, placing correct tokens among the top candidates earlier in the decoding process. This improved ranking behavior is particularly critical for morphologically rich languages where token level disambiguation requires integration of multiple linguistic features, and suggests that Indic-TunedLens better captures the hierarchical semantic refinement that occurs across transformer layers for Indian languages. Detailed rank distribution plots for each language are available in the appendix \ref{da: avg_rank}.}

\begin{figure*}
    \centering
    \includegraphics[width=0.90\linewidth]{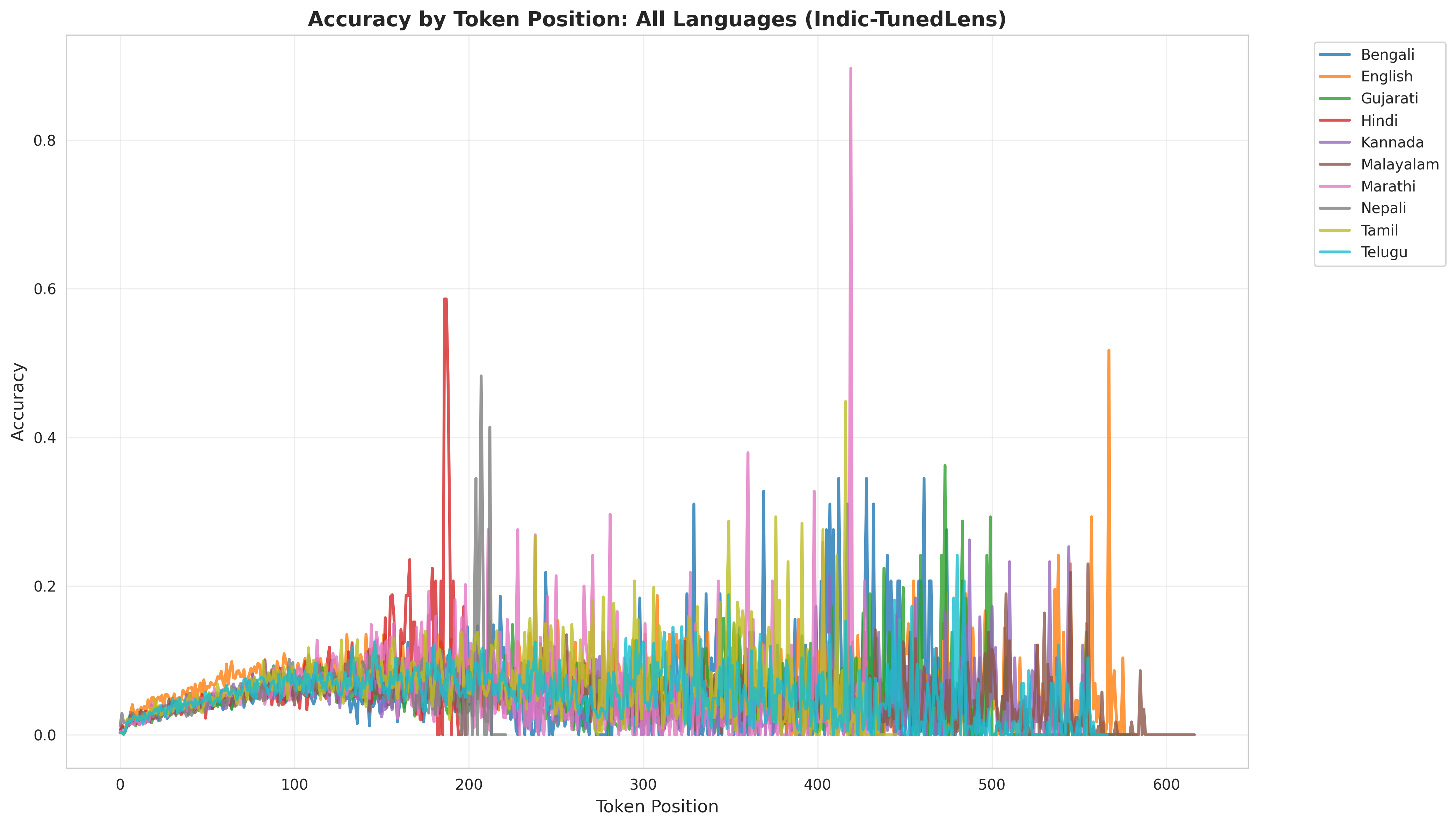}
    \caption{Accuracy by Token Positions}
    \label{fig:acc_token}
\end{figure*}

\noindent \textbf{Accuracy varies substantially across token positions, with specific positions showing language dependent spikes corresponding to answer tokens.} Figure \ref{fig:acc_token} demonstrates highly variable accuracy patterns across the 600+ token positions in the MMLU dataset samples. Marathi exhibits the most dramatic spike at position 380 (0.95 accuracy), followed by English at position 550 (0.52 accuracy) and Hindi at position 210 (0.58 accuracy). These spikes correspond to answer choice positions in the multiple choice questions, where the model's predictions align with the correct options. Most languages maintain baseline accuracy of 0.05-0.15 across question and context tokens (positions 0-400), with gradual increases toward answer regions. Bengali, Telugu, and Gujarati show multiple moderate peaks (0.20-0.35) distributed across positions 350-500, suggesting consistent prediction quality across answer choices. {The position specific accuracy patterns reveal that Indic-TunedLens successfully captures the model's reasoning trajectory through question answering tasks, with high accuracy at answer token positions indicating effective alignment between intermediate representations and final decision making processes. The language specific variation in peak positions reflects differences in tokenization schemes and morphological complexity, where languages with more complex morphology (Tamil, Telugu) require more tokens to represent equivalent semantic content, shifting the spatial distribution of critical reasoning steps. This position based analysis provides evidence that the interpretability framework maintains fidelity to the model's internal decision boundaries across diverse linguistic structures and question formats.}

\section{Discussion}

Our work establishes Indic-TunedLens as a specialized interpretability framework for morphologically rich Indian languages through three key findings from our comparative analysis.

As demonstrated in Figure \ref{fig:layer_comp}, \textit{Indic-TunedLens} enables meaningful interpretation from layer 1 onwards, whereas standard Logit Lens \cite{nostalgebraist2020logitlens} fails to capture intermediate semantics until layer 20. This 20-layer interpretability gap reveals that English centric methods fundamentally misalign with how multilingual models process morphologically rich languages. The immediate accessibility in early layers suggests that learned affine transformations successfully bridge the representational mismatch between intermediate hidden states and final vocabulary distributions. Recent work shows that language specific neurons concentrate in mid-to-late layers \cite{tang2024language,chou2025causal}, yet our framework exposes semantically meaningful states much earlier, indicating that the computational bottleneck lies in projection mechanisms rather than internal processing \cite{schut2025multilingual}.

These findings have broader implications for multilingual NLP. First, interpretability cannot be language agnostic. The same technique yields fundamentally different results depending on linguistic structure \cite{wen2023hyperpolyglot}. Second, our framework provides a methodological template for other morphologically rich families. Third, the Indic-TunedLens interpretability tool enables the identification of language specific computational regions and layers, revealing that multilingual competence arises from localized mechanisms that can be selectively analyzed or manipulated for a more controlled, fair, and reliable multilingual behavior \cite{tang2024language}. 
By introducing the Indic-TunedLens framework, we enable researchers in multilingual NLP to analyze multilingual large language models across the languages on which they are trained. By enabling understanding of how models process diverse languages, \textit{Indic-TunedLens} supports more equitable and responsible AI development in linguistically diverse regions.

\section{Conclusion}

In this paper, we have introduced Indic-TunedLens, a multilingual interpretability framework that extends the Tuned Lens paradigm to structurally diverse Indian languages. By learning language affine transformations, our approach significantly improves the layer-wise interpretability of multilingual LLMs like Sarvam-1 on Bengali, English, Gujarati, Hindi, Kannada, Malayalam,  Marathi, Nepali, Tamil and Telugu. Extensive experiments demonstrate that Indic-TunedLens outperforms the standard English centric methods in rank based accuracy, layer wise accuracy performance, and entropy alignment. Additionally, we aim to extend interpretability methods to directly improve downstream performance in multilingual tasks.

\section*{Limitations}

While \textit{Indic-TunedLens} enhances interpretability for Indian languages, it has certain limitations. Due to resource constraints and unavailability of pretraining data, we initially fine tune only the 1B-parameter Sarvam-1 model. Future work can explore multilingual models with higher number of parameters. Second, our evaluation is limited to the MMLU dataset, which restricts the diversity of tasks used to assess large language model behavior. Finally, while we achieve better interpretability, this does not directly translate to improved downstream task performance, which remains an open question for future research.

\section*{Ethics Statement}
We use publicly accessible datasets for our experiments, strictly for academic purposes and in full accordance with their licensing terms.

\section*{Acknowledgments}
Mamta gratefully acknowledges the support from the Engineering and Physical Sciences Research Council (EPSRC, grant number EP/X04162X/1).


\bibliography{custom}

\appendix


\section{Experimental Setup} \label{experiment_setup}

\subsection{Dataset}
\label{appendix:sangraha}

\paragraph{Sangraha Dataset}

The Sangraha dataset serves as our primary training corpus for developing language specific interpretability lenses that can effectively analyze the internal representations of multilingual models. As detailed in Table A.1, the composition of the dataset is strategically balanced to ensure adequate representation in all languages while maintaining computational feasibility.

\begin{table}[h!]
\centering
\begin{tabular}{l r}
\hline
\textbf{Language} & \textbf{Number of Rows} \\
\hline
Bengali   &   11497021 \\
English   &   17482249 \\
Gujarati  &   3970097 \\
Hindi     &   17420932 \\
Kannada   &   3632345 \\
Malayalam &   6370342 \\
Marathi   &   5865617 \\
Panjabi   &   1738597 \\
Nepali    &   91476306 \\
Tamil     &   7828512 \\
Telugu    &   7081734 \\
\hline
\end{tabular}
\caption{Sangraha Dataset for Training Sarvam Tuned Lens}
\label{tab:sangraha_tl}
\end{table}

The substantial size of each language subset ensures sufficient statistical power for training robust interpretability mechanisms, while the linguistic diversity across multiple Indian languages provides the necessary foundation for developing generalizable interpretability methods.

\paragraph{MMLU Dataset}

Our evaluation framework employs a curated subset of the multilingual Massive Multitask Language Understanding (MMLU) dataset, adapted for Indian languages \citet{dac2023okapi}. As shown in Table~\ref{tab:mmlu_yl}, the testing dataset includes samples across ten languages, with relatively balanced distribution ranging from 215 to 277 samples per language. This balanced distribution enables fair comparative analysis while testing the generalization capabilities of our approach across the diverse linguistic landscape.

\begin{table}[h!]
\centering
\begin{tabular}{l r}
\hline
\textbf{Language} & \textbf{Number of Rows} \\
\hline
Bengali    &  216 \\
English    &  277 \\
Gujarati   &  243 \\
Hindi      &  235 \\
Kannada    &  261 \\
Malayalam  &  265 \\
Marathi    &  221 \\
Nepali     &  215 \\
Tamil      &  251 \\
Telugu     &  272 \\
\hline
\end{tabular}
\caption{MMLU Dataset for Testing Tuned Lens}
\label{tab:mmlu_yl}
\end{table}

\subsection{Implementation Details}\label{appendix:implementationdetails}

Our experimental framework is built around the Sarvam-1 model, which has been specifically optimized for 10 Indian languages: Bengali (bn), English (en) Gujarati (gu), Hindi (hi), Kannada (kn), Malayalam (ml), Marathi (mr), Nepali(np), Punjabi (pa), Tamil (ta), and Telugu (te). This multilingual optimization makes it an ideal candidate for studying interpretability across diverse Indian languages, as it inherently possesses cross lingual capabilities developed through extensive multilingual pre training.

For training the tuned lens, we utilize the \texttt{tuned-lens} package, which provides a robust framework for training and evaluating lenses on transformer models like Sarvam-1 \cite{belrose2023eliciting}. The package enables us to probe intermediate layers of the model to analyze how multilingual tokens influence prediction patterns.

\begin{table}[h!]
\centering
\begin{tabular}{l r}
\hline
\textbf{Hyperparameter} & \textbf{Value} \\
\hline
Model Name & sarvamai/sarvam-1 \\
\# of Nodes & 1 \\
Processes per Node & 5 \\
Per-GPU Batch Size & 1 \\
FSDP & Enabled \\
Launch Mode & Standalone \\
\hline
\end{tabular}
\caption{Training Configuration and Hyperparameters for Tuned Lens with Sarvam-1}
\label{tab:tl_hyperparams}
\end{table}

The training configuration detailed in Table~\ref{tab:tl_hyperparams} utilizes a distributed setup with careful resource allocation to ensure stable training while maintaining computational efficiency. We employ Fully Sharded Data Parallel (FSDP) training to handle the large scale nature of the Sarvam-1 model effectively. The single node configuration with 5 processes per node represents an optimal balance between computational efficiency and resource availability. The per GPU batch size of 1 is chosen to maximize memory utilization while preventing out of memory errors during training of the tuned lens components, given the substantial memory requirements of the Sarvam-1 model architecture.



\section{Detailed Analysis}

\subsection{Language Wise Layer-wise Accuracy Comparison between Indic-TunedLens and LogitLens} \label{da: language_acc}

\begin{figure}[htb!]
    \centering
    \includegraphics[width=1\linewidth]{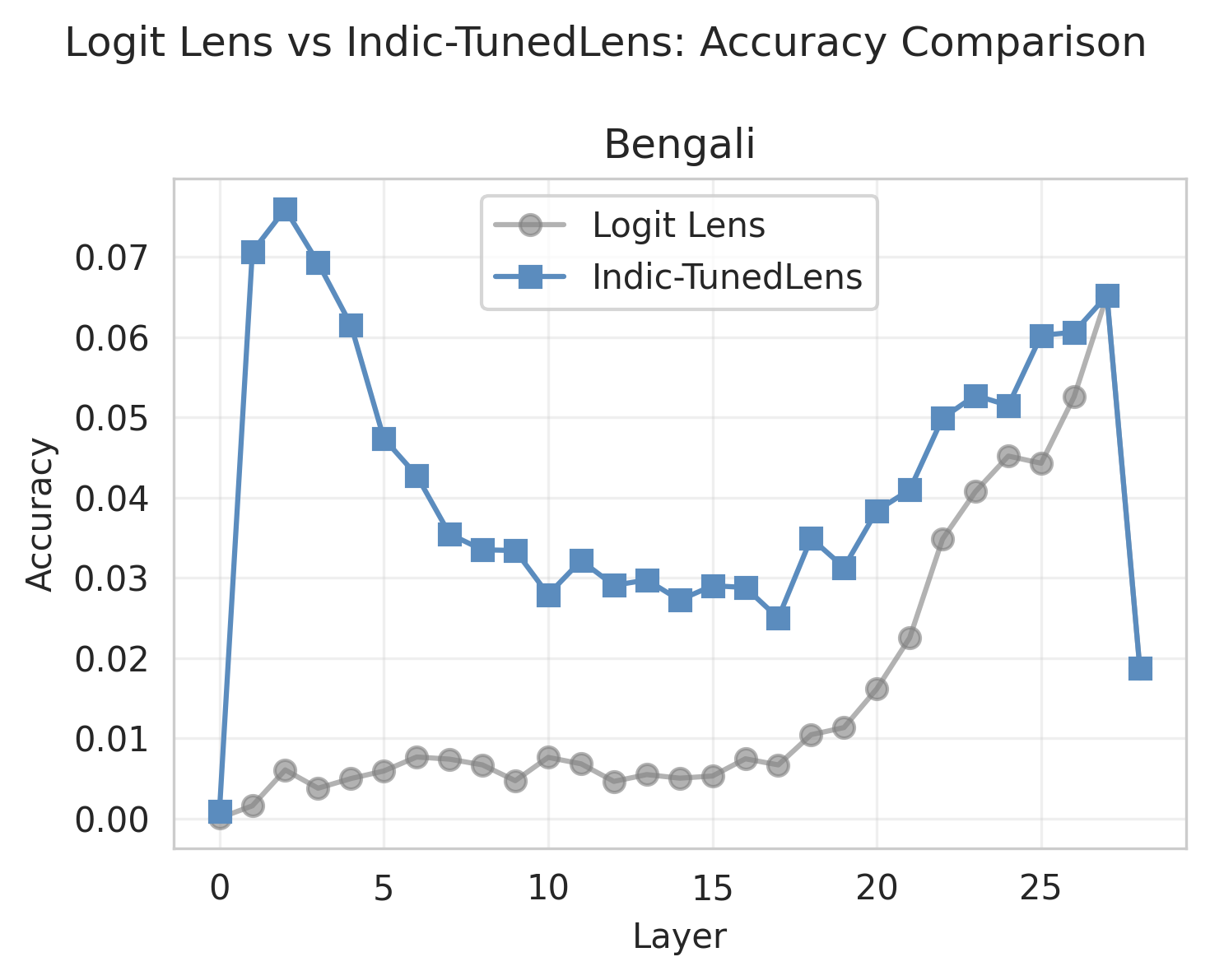}
    \caption{Layer-wise Accuracy Comparison for Bengali}
    \label{layer_bn}
\end{figure}

\begin{figure}[htb!]
    \centering
    \includegraphics[width=1\linewidth]{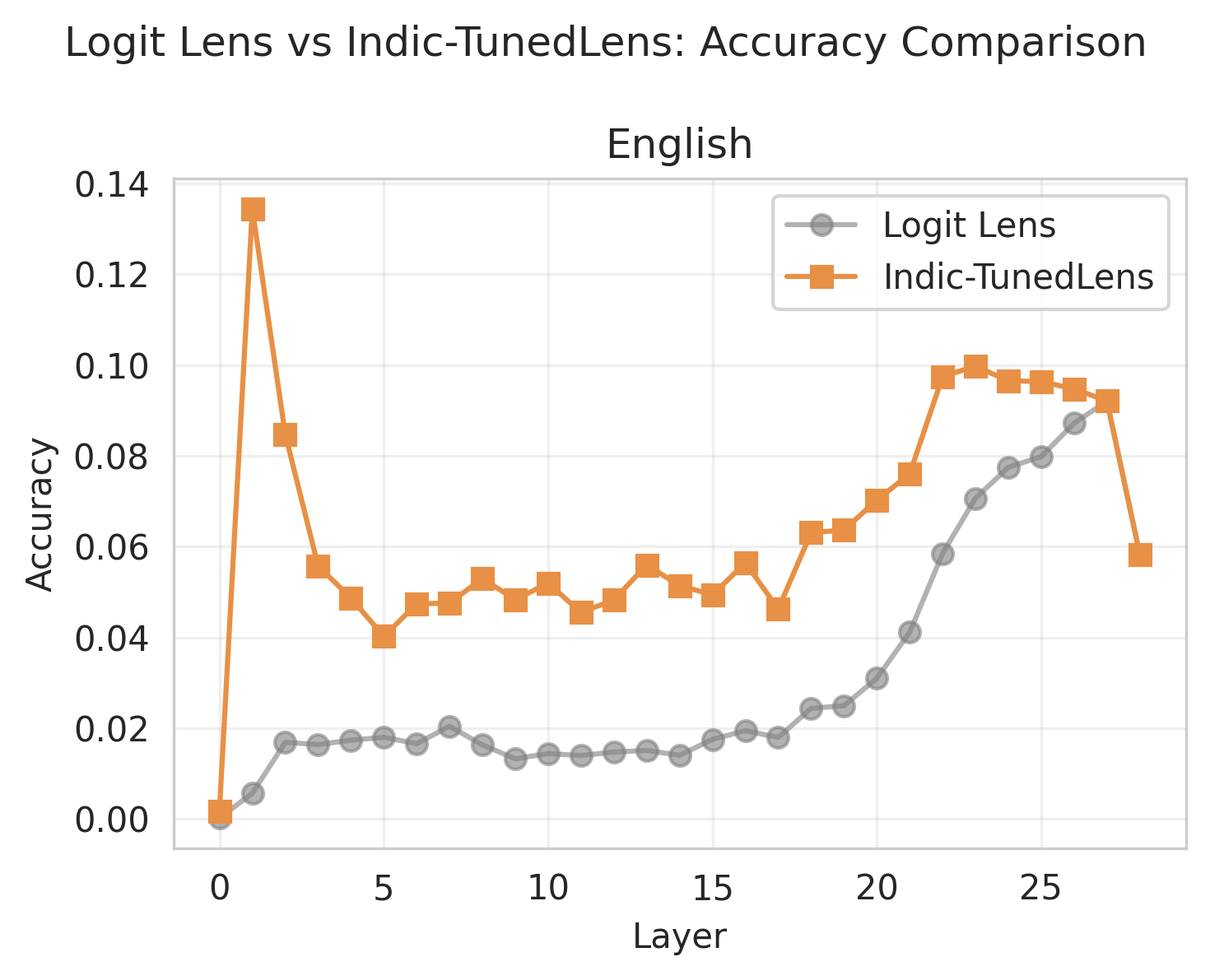}
    \caption{Layer-wise Accuracy Comparison for English}
    \label{layer_en}
\end{figure}

\begin{figure}[htb!]
    \centering
    \includegraphics[width=1\linewidth]{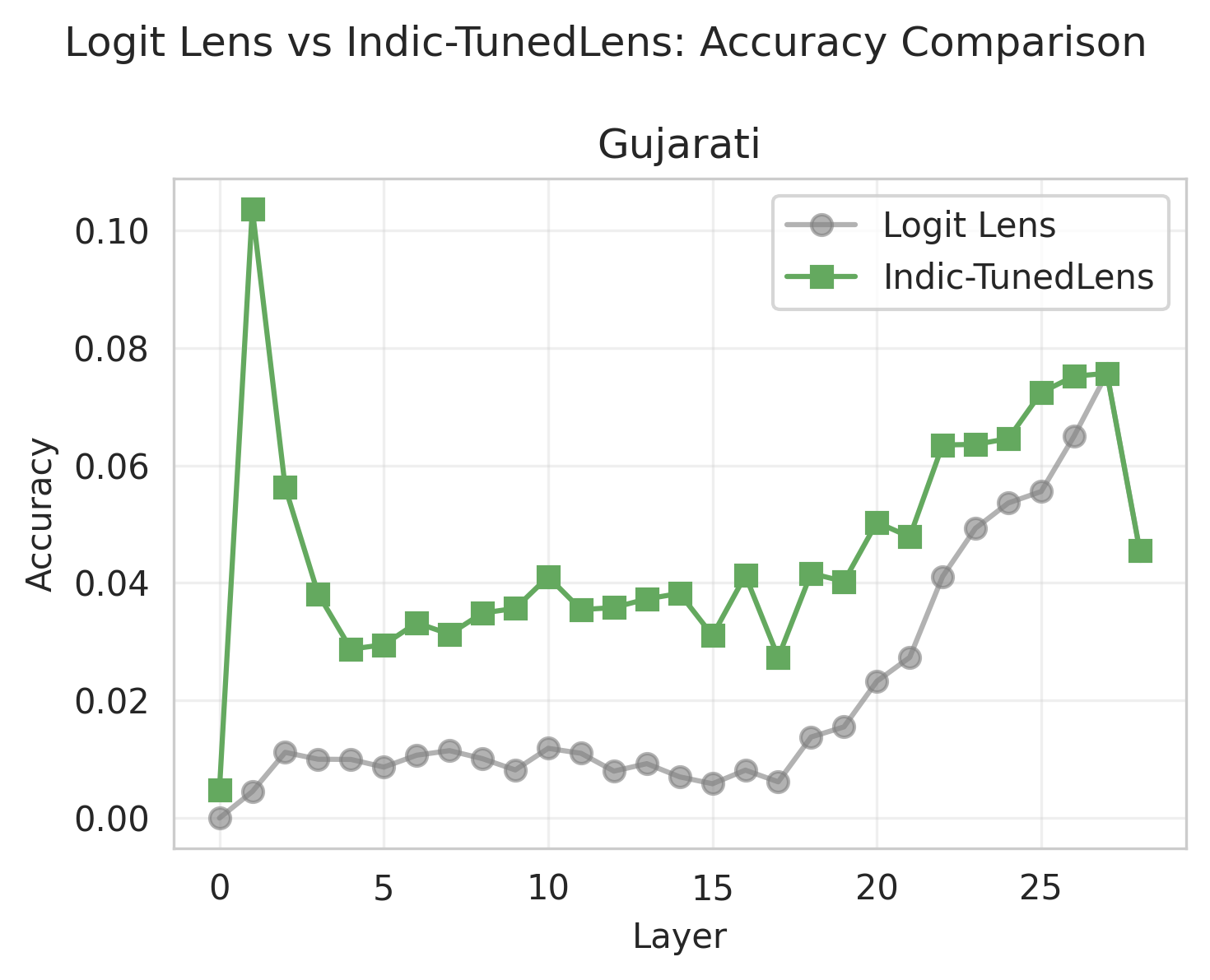}
    \caption{Layer-wise Accuracy Comparison for Gujarati}
    \label{layer_gu}    
\end{figure}

\begin{figure}[htb!]
    \centering
    \includegraphics[width=1\linewidth]{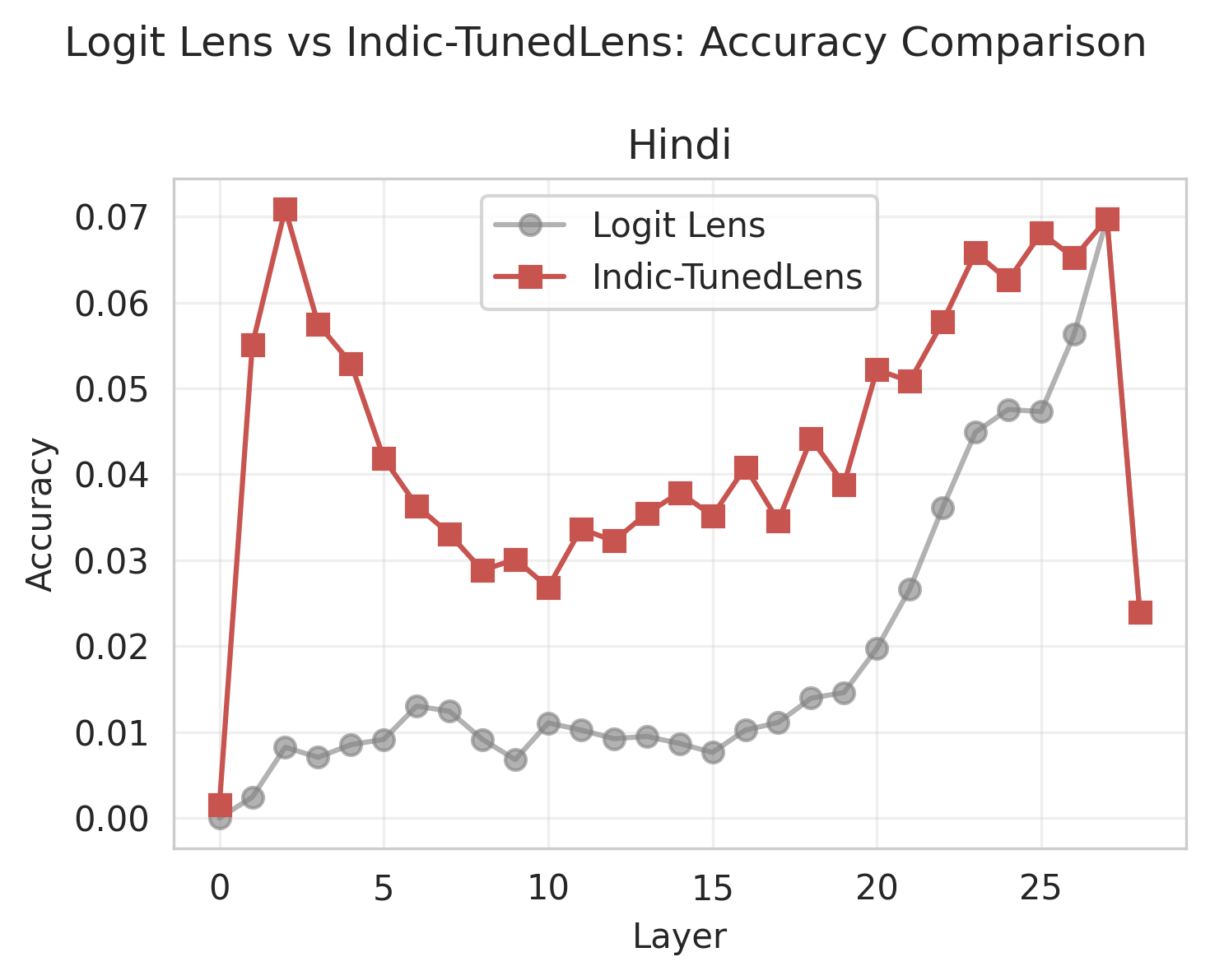}
    \caption{Layer-wise Accuracy Comparison for Hindi}
    \label{layer_hi}    
\end{figure}

\begin{figure}[htb!]
    \centering
    \includegraphics[width=1\linewidth]{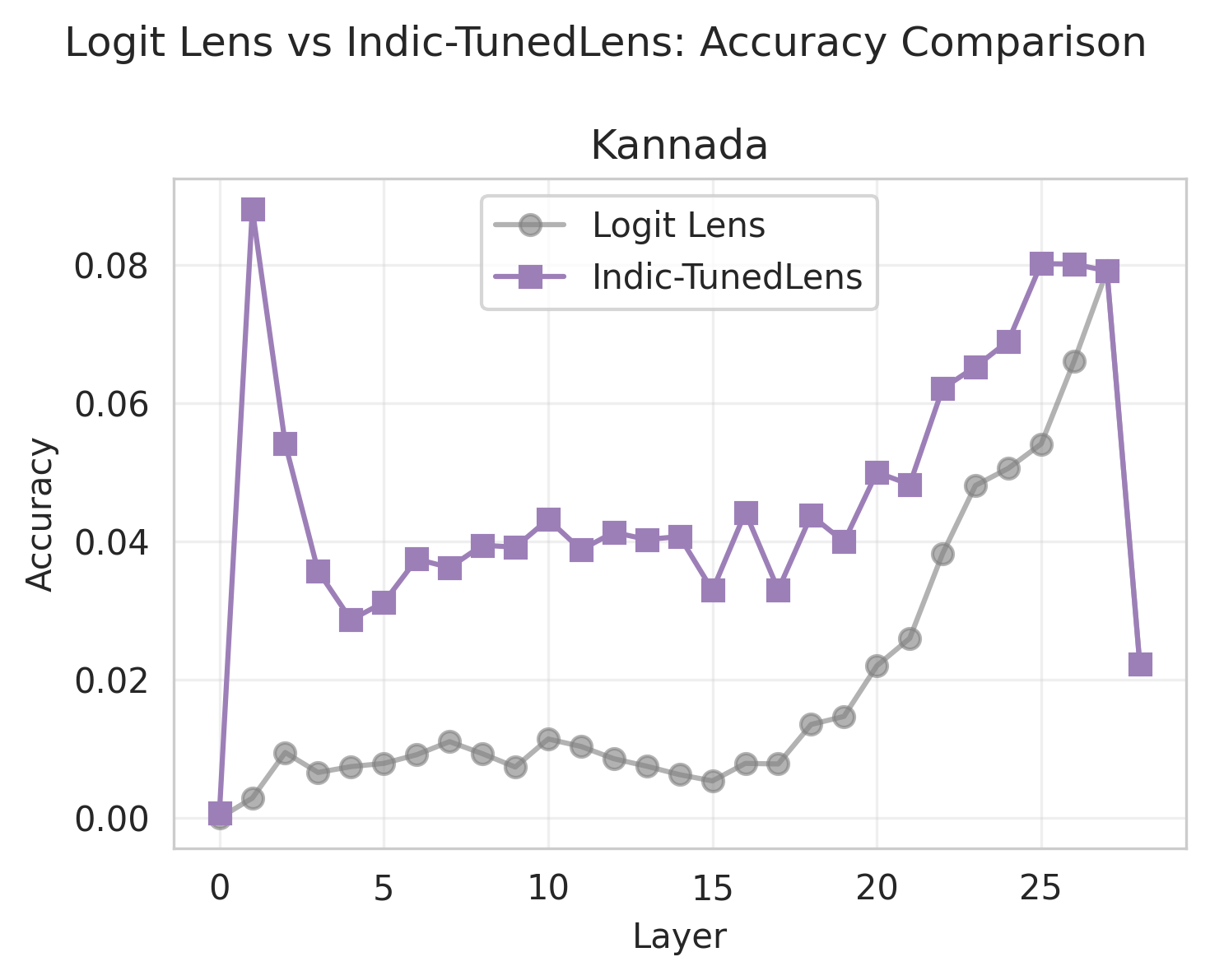}
    \caption{Layer-wise Accuracy Comparison for Kannada}
    \label{layer_kn}    
\end{figure}

\begin{figure}[htb!]
    \centering
    \includegraphics[width=1\linewidth]{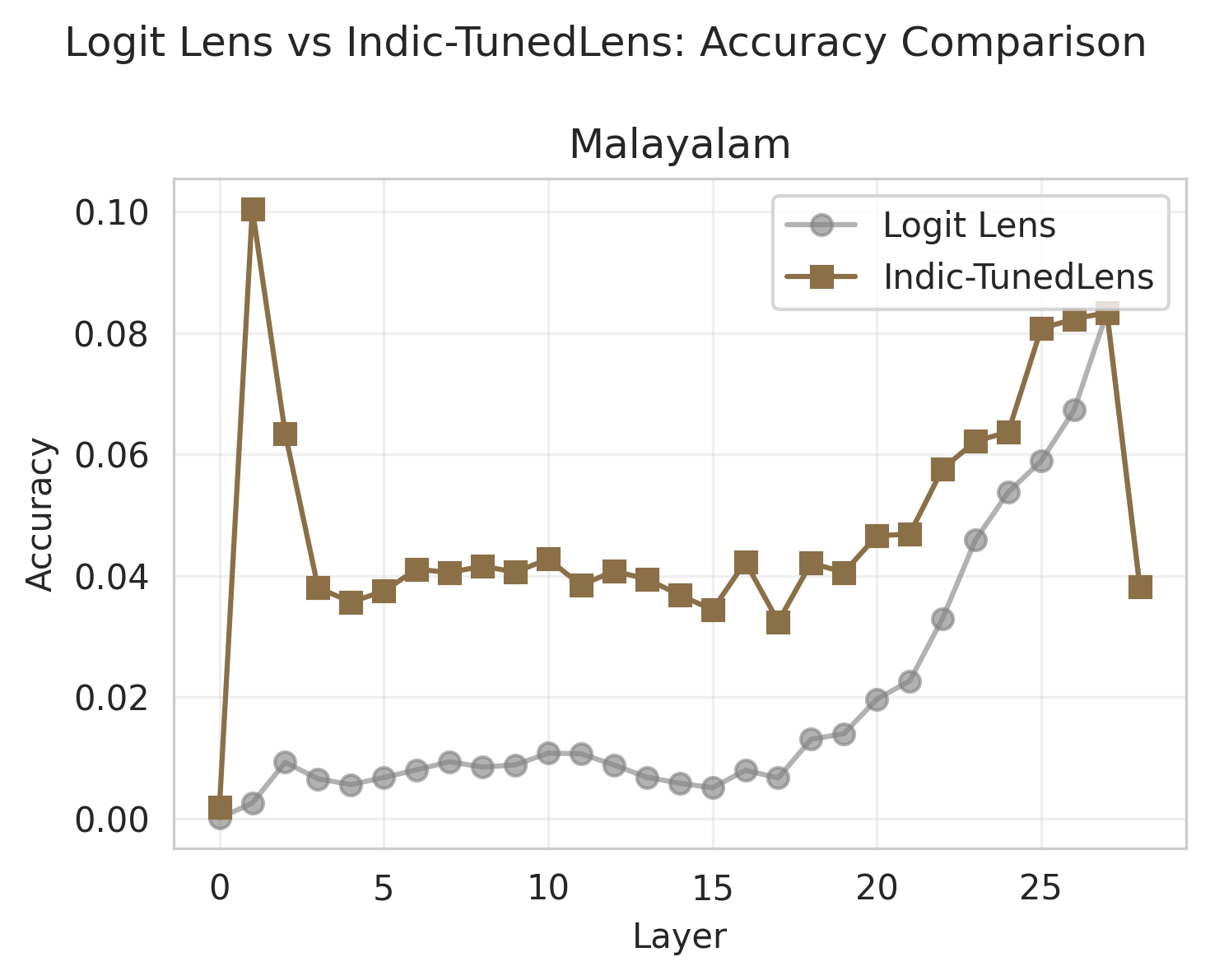}
    \caption{Layer-wise Accuracy Comparison for Malayalam}
    \label{layer_ml}    
\end{figure}

\begin{figure}[htb!]
    \centering
    \includegraphics[width=1\linewidth]{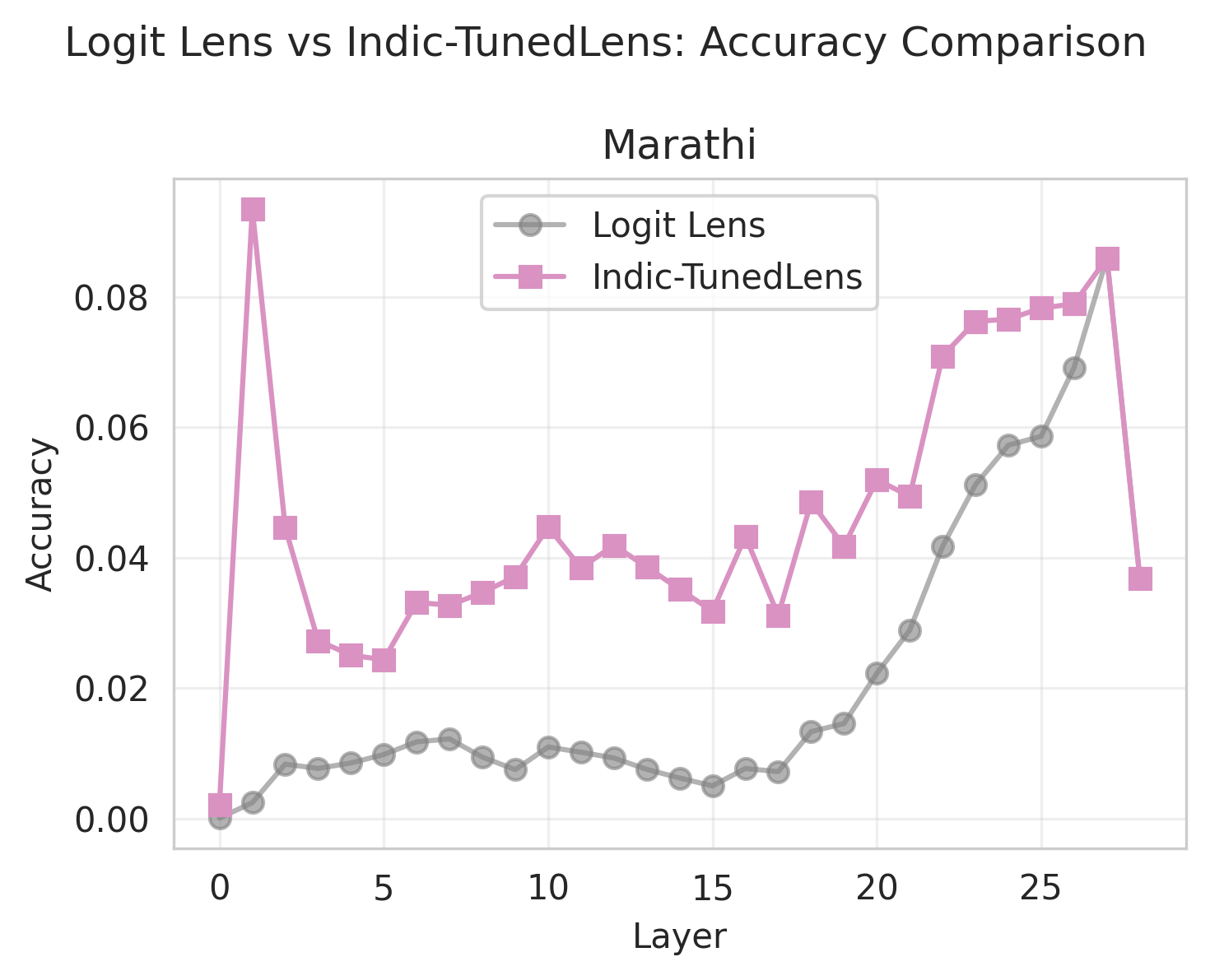}
    \caption{Layer-wise Accuracy Comparison for Marathi}
    \label{layer_mr}    
\end{figure}

\begin{figure}[htb!]
    \centering
    \includegraphics[width=1\linewidth]{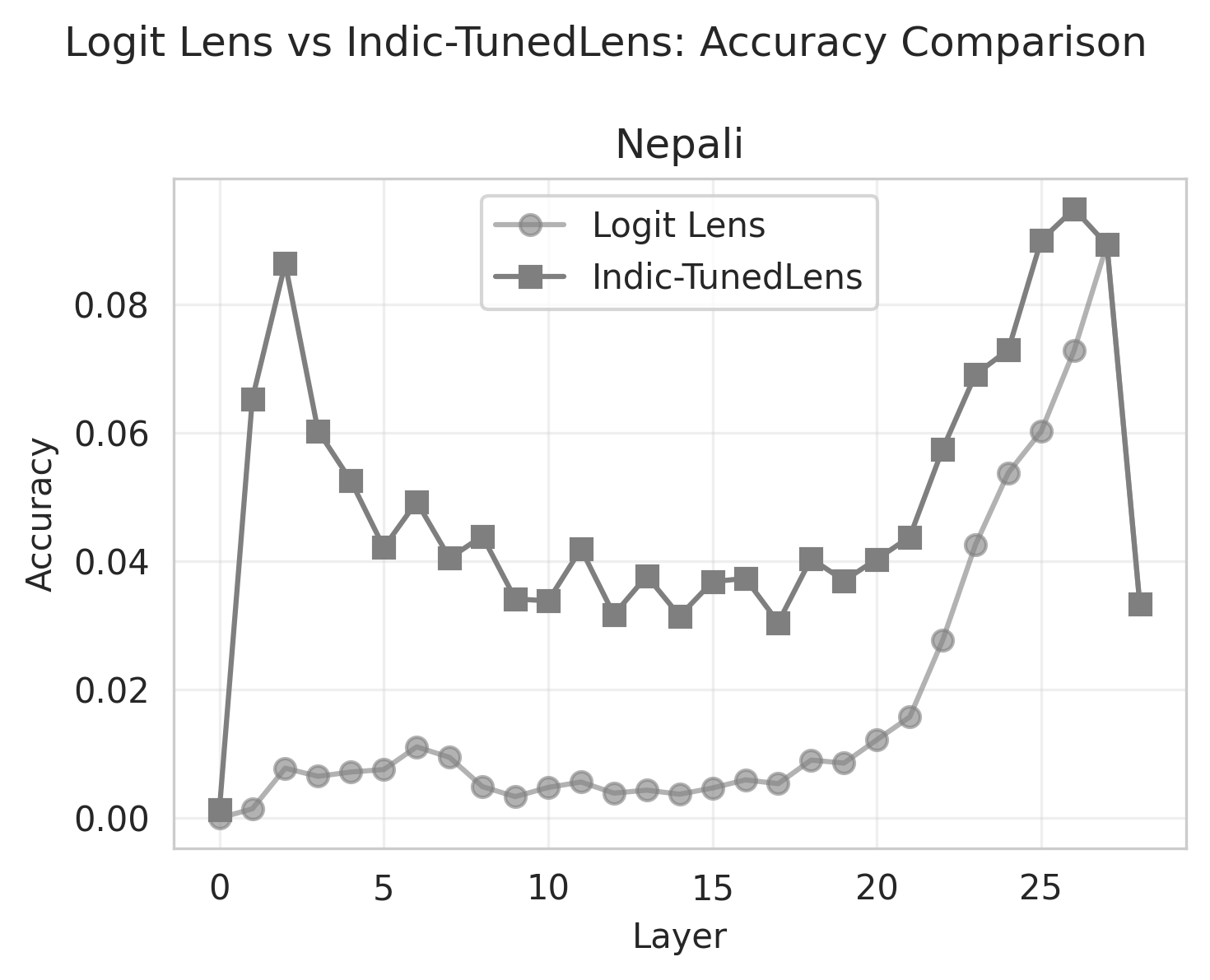}
    \caption{Layer-wise Accuracy Comparison for Nepali}
    \label{layer_np}    
\end{figure}

\begin{figure}[htb!]
    \centering
    \includegraphics[width=1\linewidth]{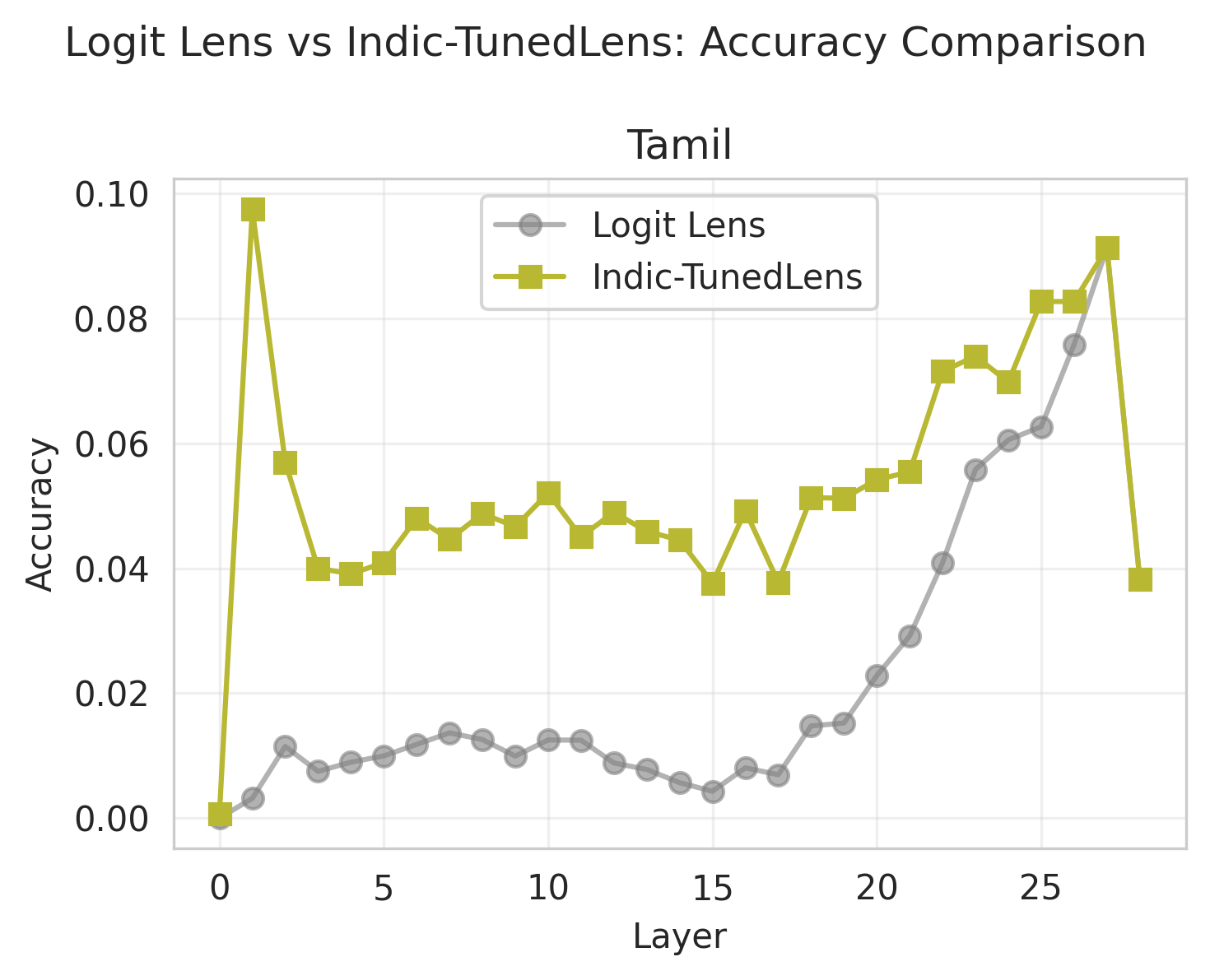}
    \caption{Layer-wise Accuracy Comparison for Tamil}
    \label{layer_ta}    
\end{figure}

\begin{figure}[htb!]
    \centering
    \includegraphics[width=1\linewidth]{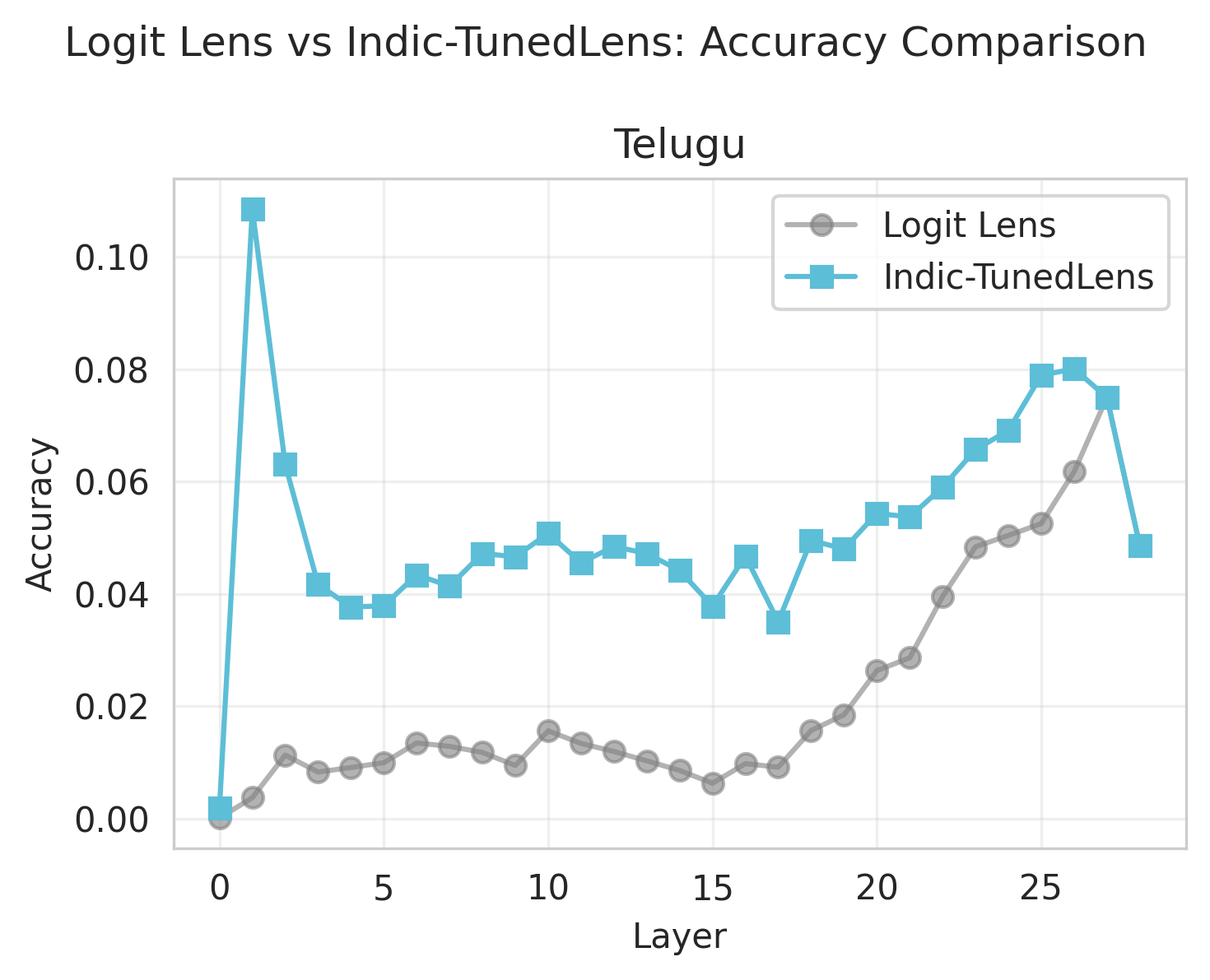}
    \caption{Layer-wise Accuracy Comparison for Telugu}
    \label{layer_te}    
\end{figure}

This section presents a comprehensive language specific analysis of layer wise accuracy patterns, comparing the performance of Indic-TunedLens against the standard Logit Lens across all evaluated Indian languages. Each plot illustrates the accuracy across all 28 layers of Sarvam-1 model, revealing distinct processing characteristics for individual languages. The layer-wise accuracy comparison is shown for Bengali in Figure~\ref{layer_bn}, English in Figure~\ref{layer_en}, Gujarati in Figure~\ref{layer_gu}, Hindi in Figure~\ref{layer_hi}, Kannada in Figure~\ref{layer_kn}, Malayalam in Figure~\ref{layer_ml}, Marathi in Figure~\ref{layer_mr}, Nepali in Figure~\ref{layer_np}, Tamil in Figure~\ref{layer_ta}, and Telugu in Figure~\ref{layer_te}.

\FloatBarrier
\subsection{Language Wise Average Rank of Correct Predictions between Indic-TunedLens and LogitLens} \label{da: avg_rank}

\begin{figure}[htb!]
    \centering
    \includegraphics[width=1\linewidth]{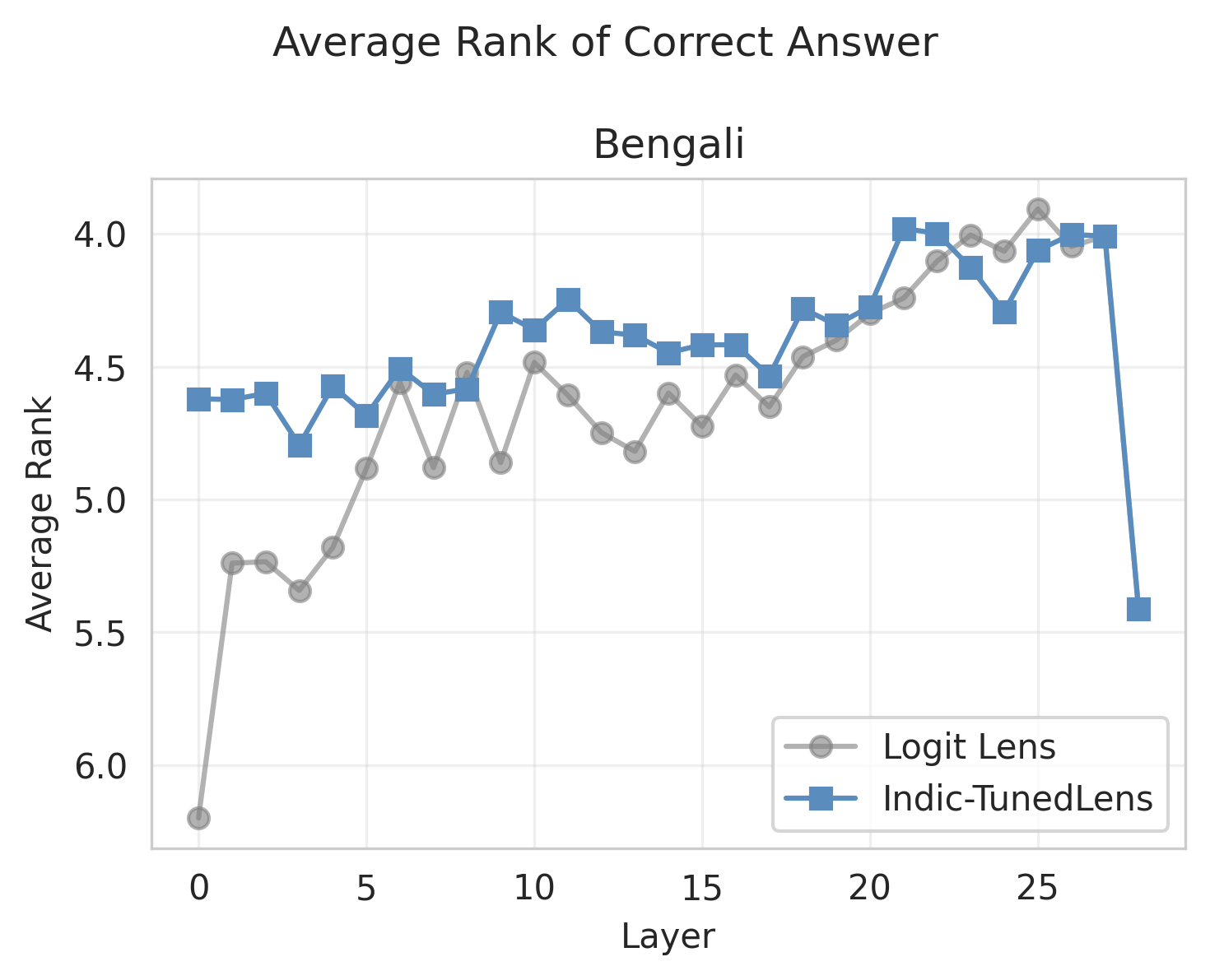}
    \caption{Average Rank of Correct Predictions for Bengali}
    \label{avg_bn}
\end{figure}

\begin{figure}[htb!]
    \centering
    \includegraphics[width=1\linewidth]{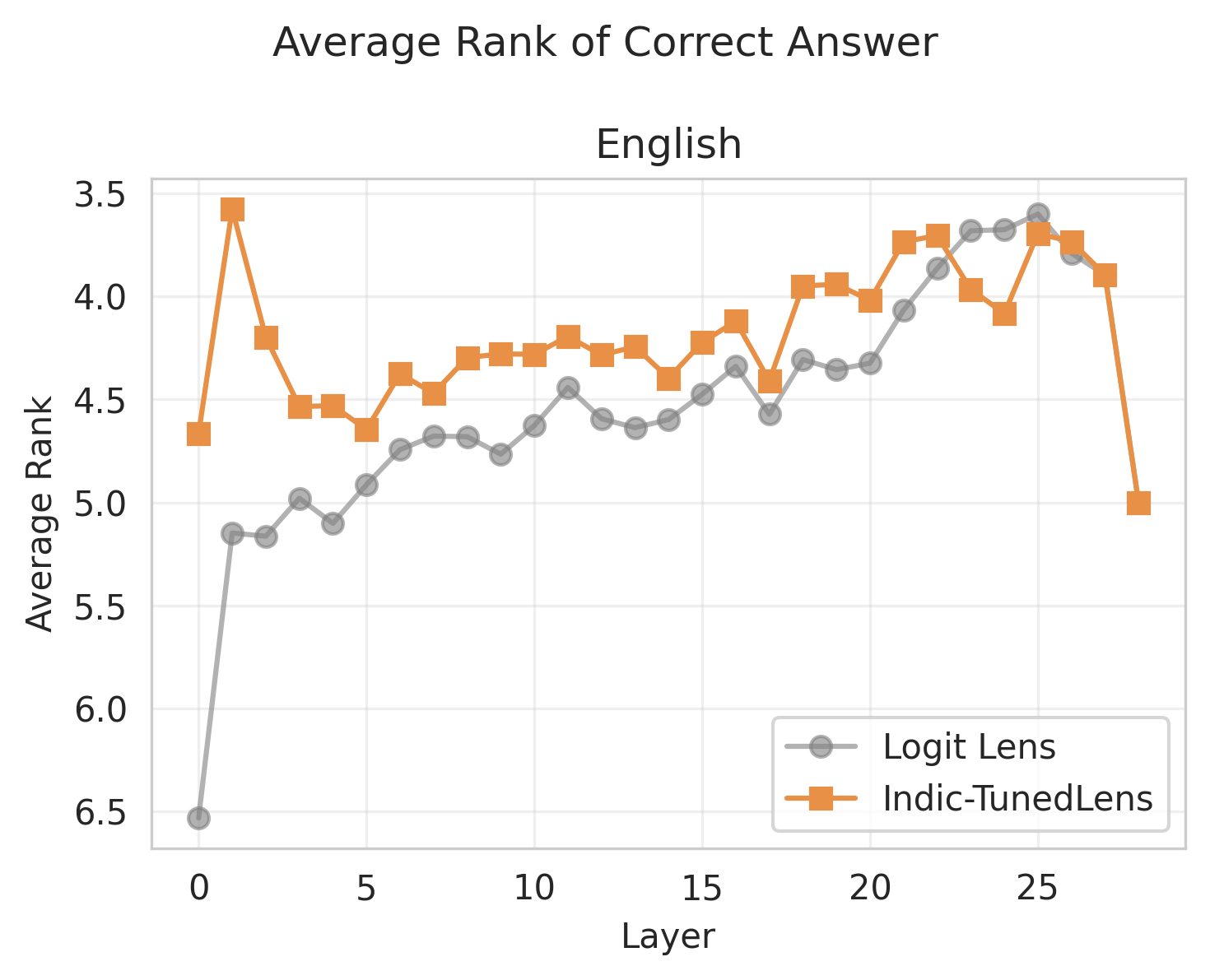}
    \caption{Average Rank of Correct Predictions for English}
    \label{avg_en}
\end{figure}

\begin{figure}[htb!]
    \centering
    \includegraphics[width=1\linewidth]{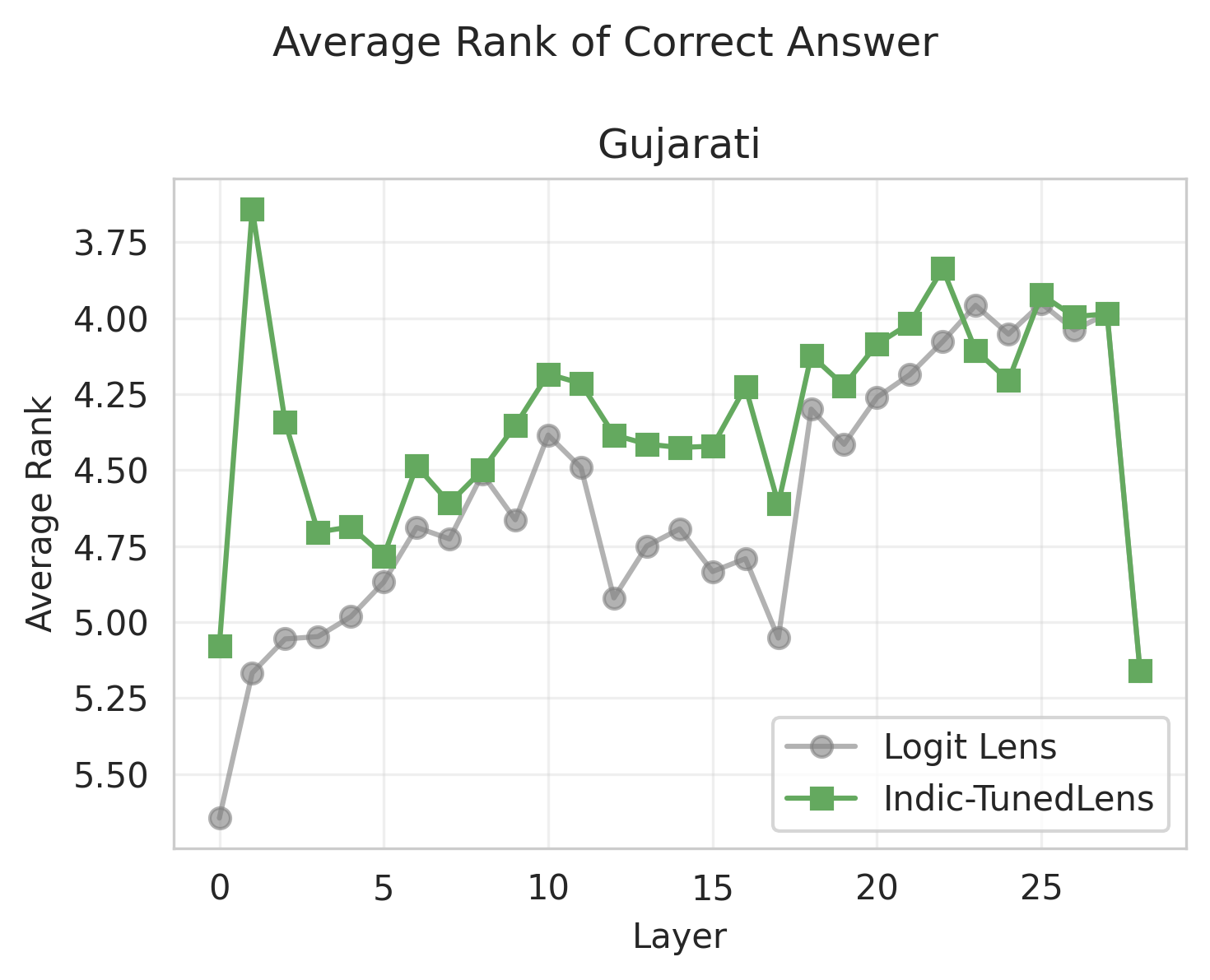}
    \caption{Average Rank of Correct Predictions for Gujarati}
    \label{avg_gu}
\end{figure}

\begin{figure}[htb!]
    \centering
    \includegraphics[width=1\linewidth]{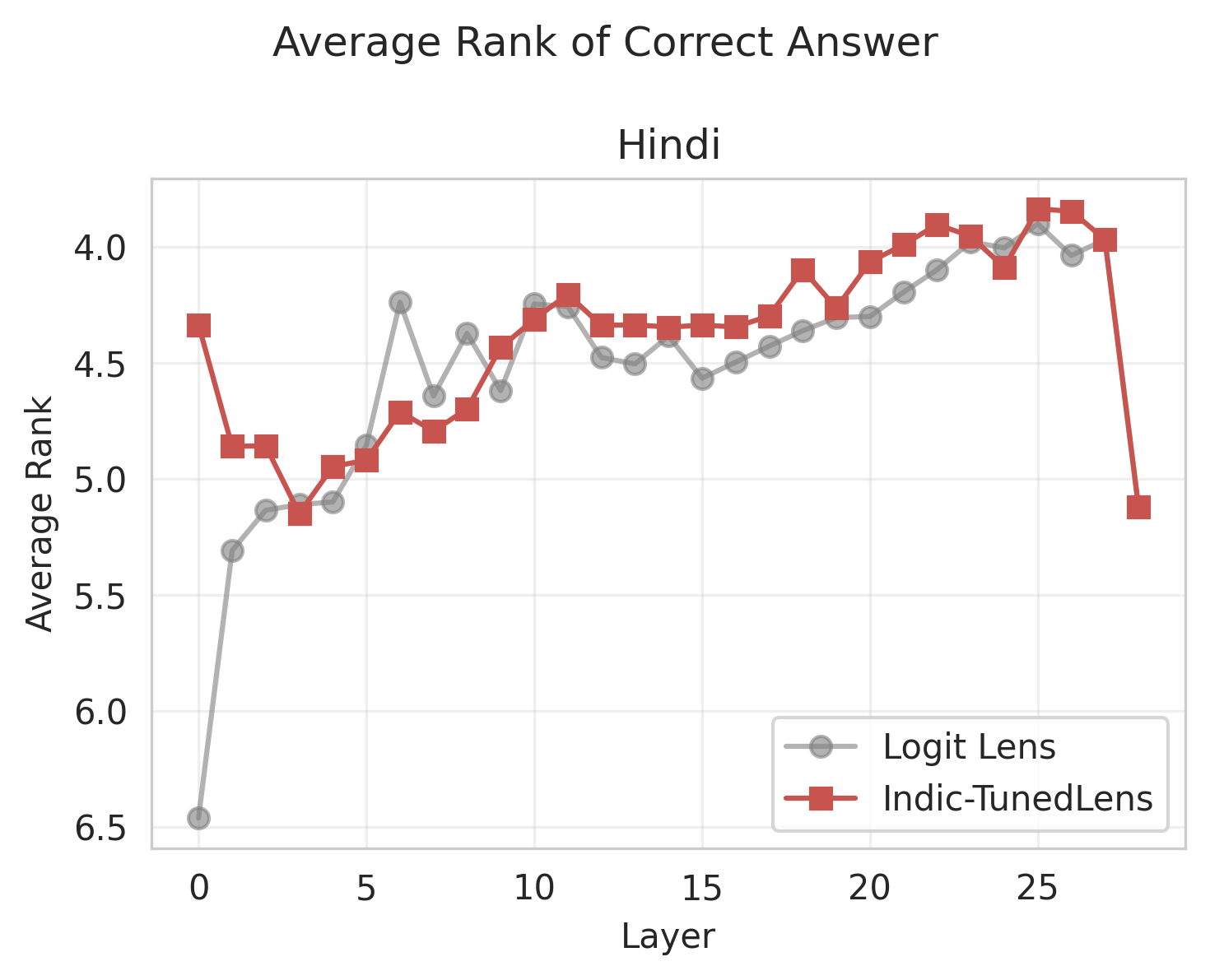}
    \caption{Average Rank of Correct Predictions for Hindi}
    \label{avg_hi}
\end{figure}

\begin{figure}[htb!]
    \centering
    \includegraphics[width=1\linewidth]{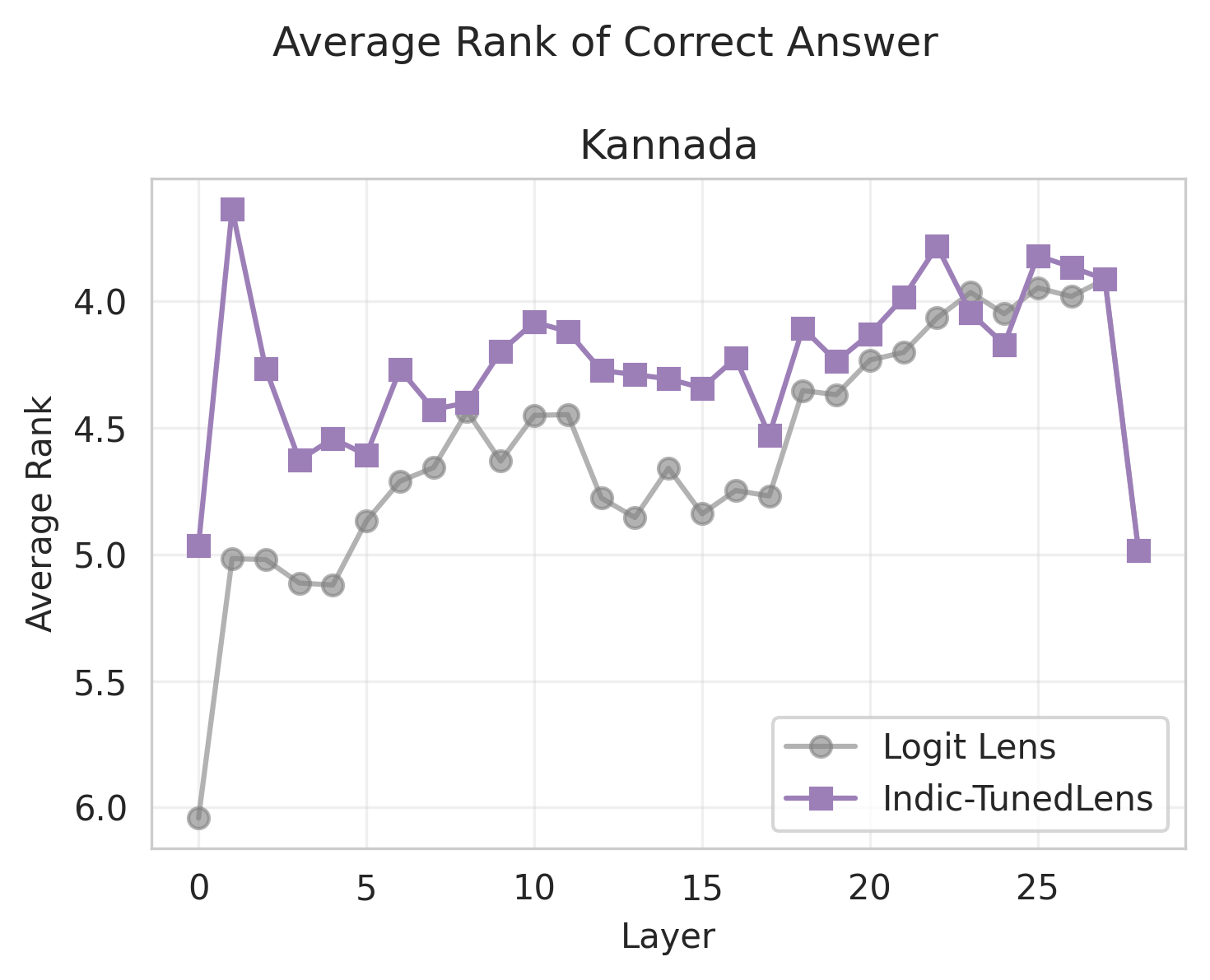}
    \caption{Average Rank of Correct Predictions for Kannada}
    \label{avg_kn}
\end{figure}

\begin{figure}[htb!]
    \centering
    \includegraphics[width=1\linewidth]{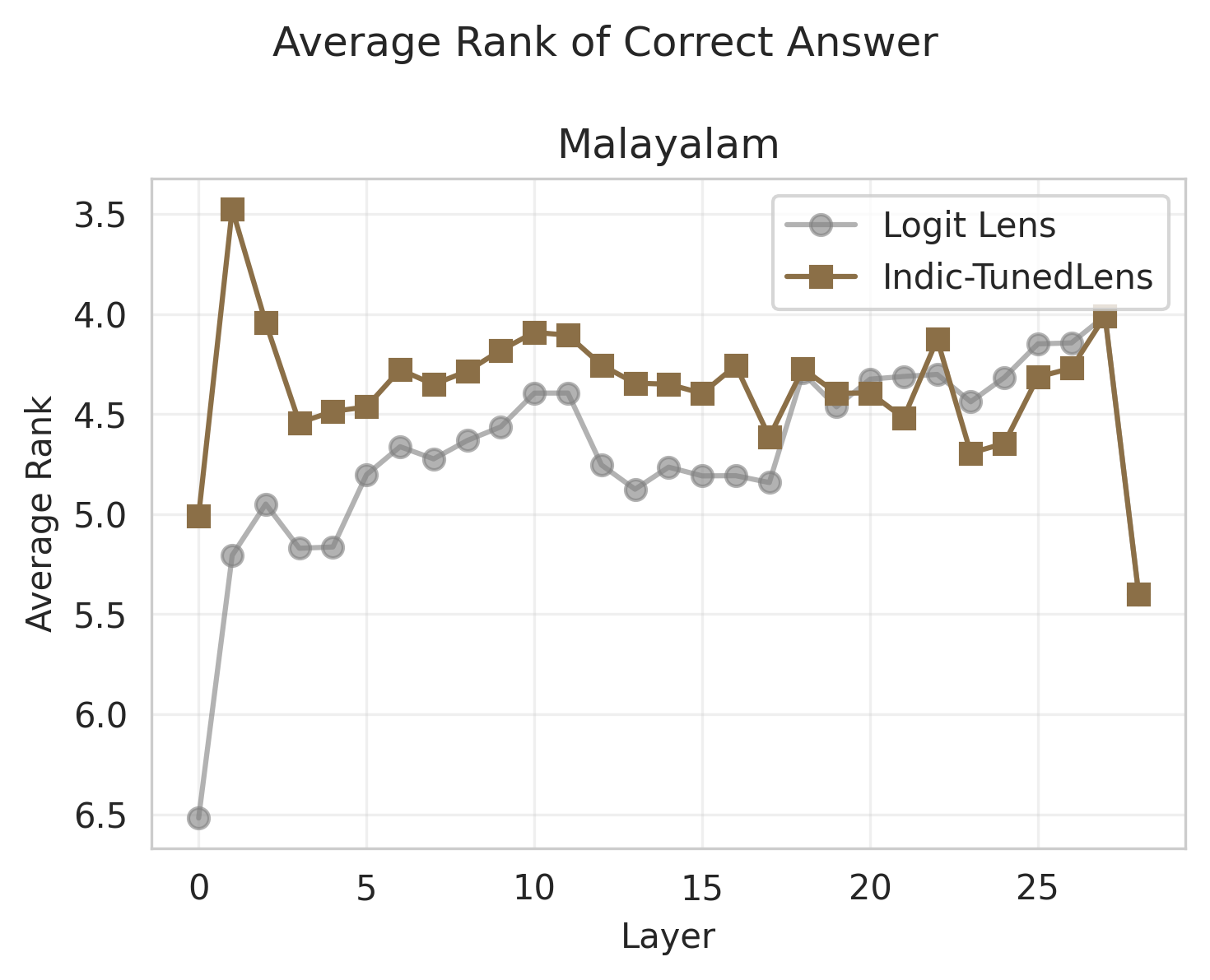}
    \caption{Average Rank of Correct Predictions for Malayalam}
    \label{avg_ml}
\end{figure}

\begin{figure}[htb!]
    \centering
    \includegraphics[width=1\linewidth]{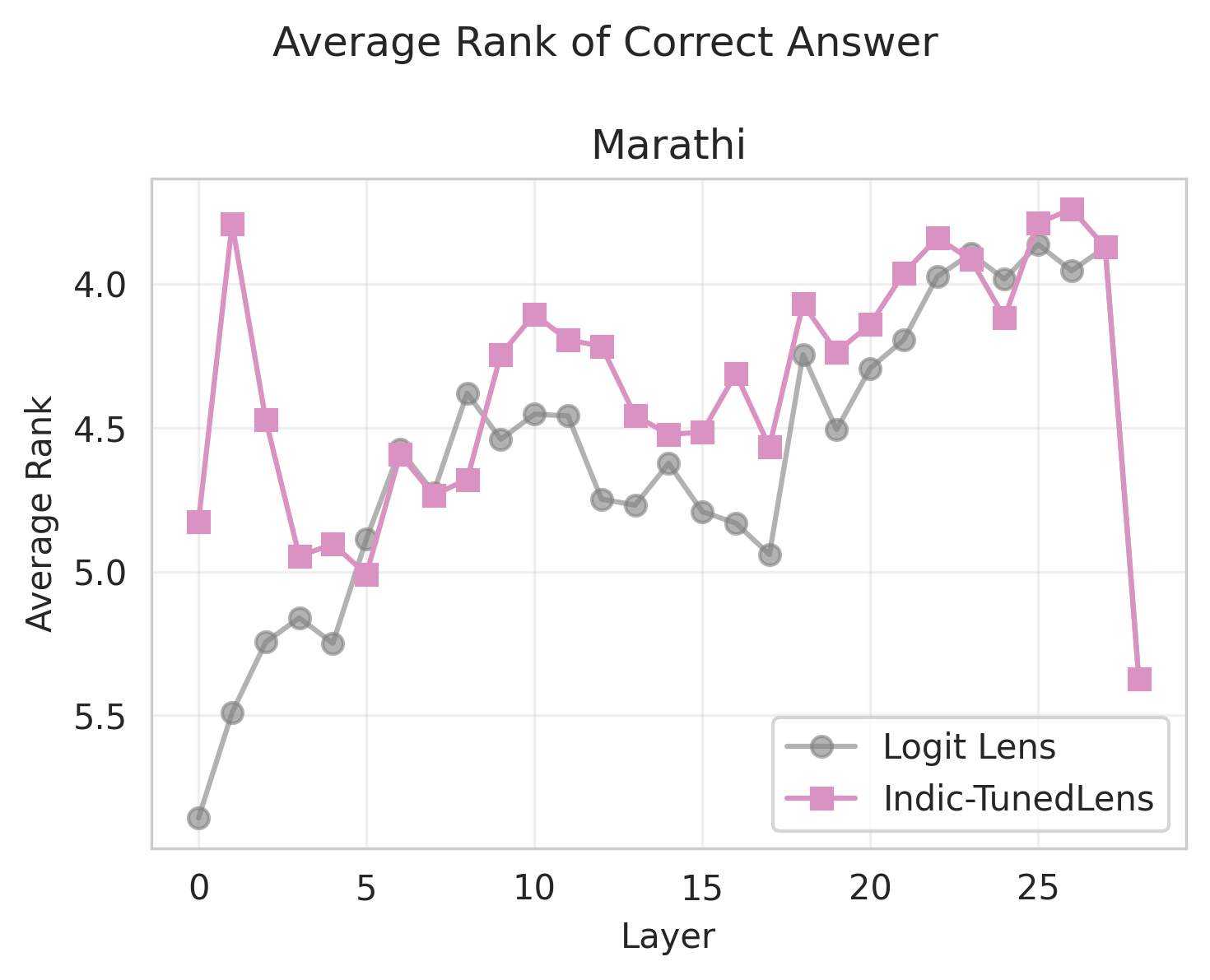}
    \caption{Average Rank of Correct Predictions for Marathi}
    \label{avg_mr}
\end{figure}

\begin{figure}[htb!]
    \centering
    \includegraphics[width=1\linewidth]{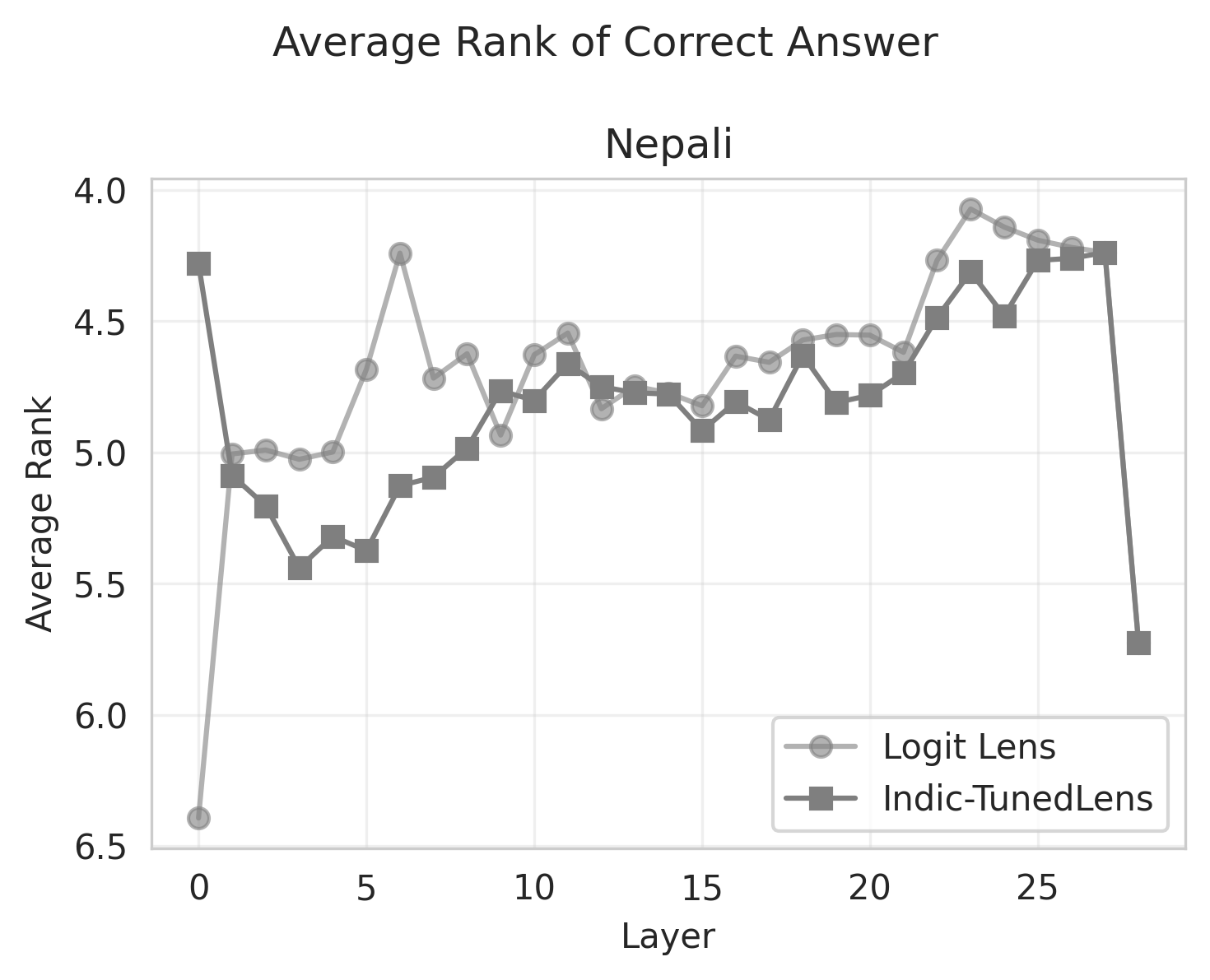}
    \caption{Average Rank of Correct Predictions for Nepali}
    \label{avg_ne}
\end{figure}

\begin{figure}[htb!]
    \centering
    \includegraphics[width=1\linewidth]{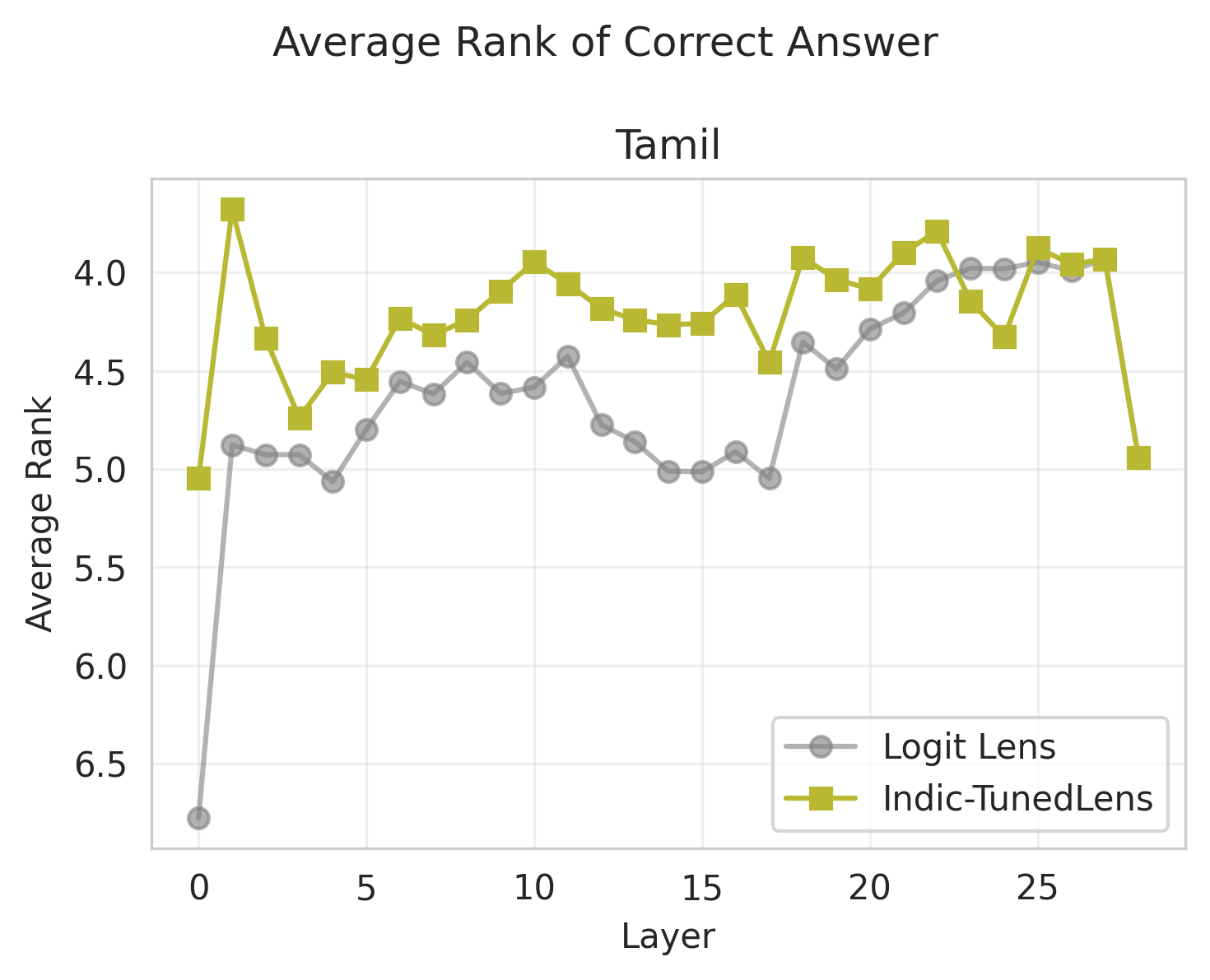}
    \caption{Average Rank of Correct Predictions for Tamil}
    \label{avg_ta}
\end{figure}

\begin{figure}[htb!]
    \centering
    \includegraphics[width=1\linewidth]{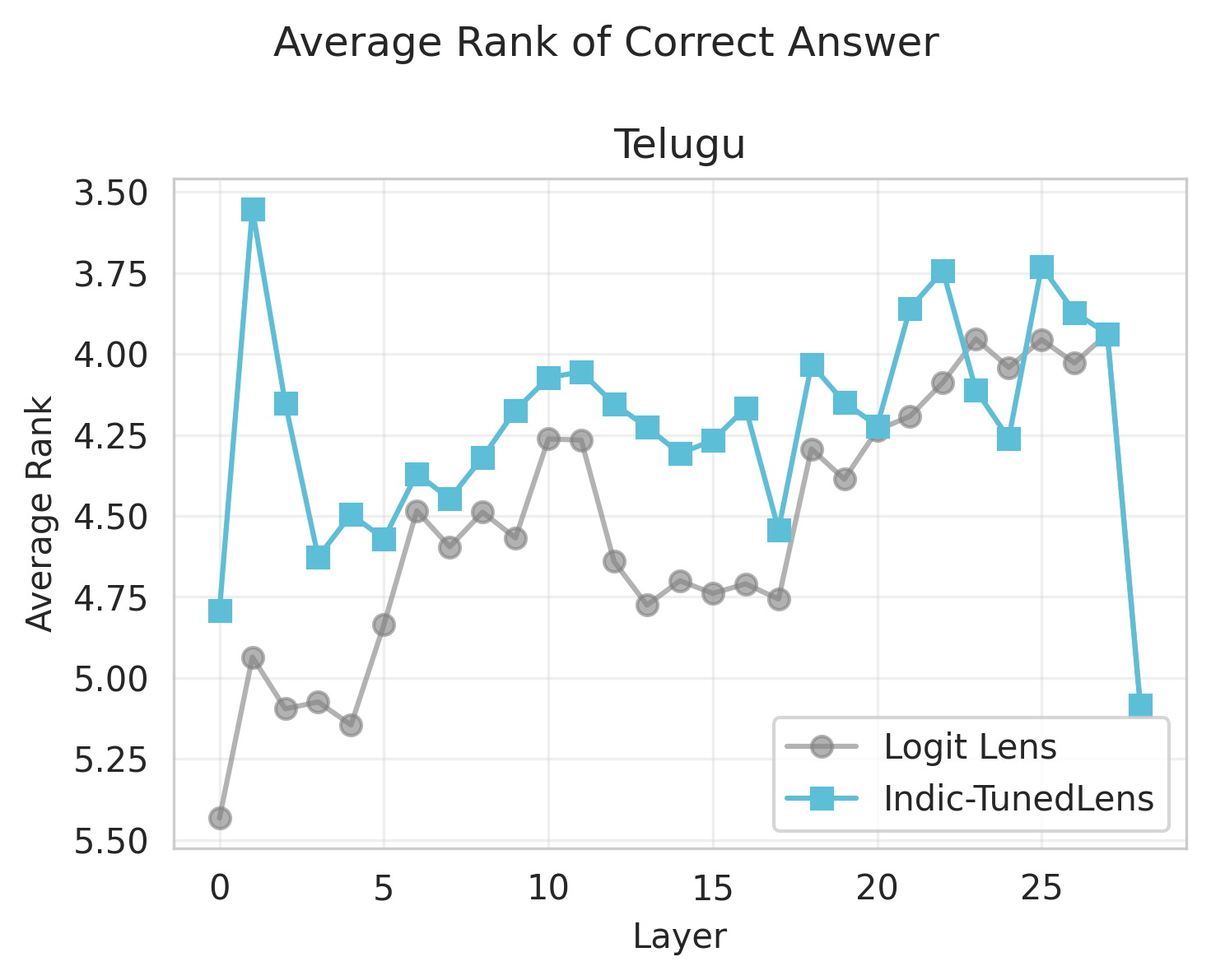}
    \caption{Average Rank of Correct Predictions for Telugu}
    \label{avg_te}
\end{figure}

This section provides a detailed examination of the average rank positions assigned to correct tokens across all layers of each language. The average rank metric quantifies how confidently the model places the correct prediction among its top candidates at each layer, with lower ranks indicating higher confidence and better alignment between intermediate representations and final outputs. The average rank of correct predictions is shown for Bengali in Figure~\ref{avg_bn}, English in Figure~\ref{avg_en}, Gujarati in Figure~\ref{avg_gu}, Hindi in Figure~\ref{avg_hi}, Kannada in Figure~\ref{avg_kn}, Malayalam in Figure~\ref{avg_ml}, Marathi in Figure~\ref{avg_mr}, Nepali in Figure~\ref{avg_ne}, Tamil in Figure~\ref{avg_ta}, and Telugu in Figure~\ref{avg_te}.

\end{document}